\RequirePackage{fix-cm}
\documentclass[smallextended]{svjour3}       
\smartqed  
\usepackage{graphicx}
\usepackage{makecell}
\usepackage{enumitem}
\usepackage{multirow}%
\usepackage{amsmath,amssymb,amsfonts}%
\usepackage[numbers,sort&compress]{natbib}
\usepackage{bm}
\usepackage{mathrsfs}%
\usepackage[title]{appendix}%
\usepackage[table]{xcolor}
\usepackage{xcolor}%
\usepackage{textcomp}%
\usepackage{manyfoot}%
\usepackage{booktabs}%
\usepackage{algorithm}%
\usepackage{algorithmicx}%
\usepackage{algpseudocode}%
\usepackage{listings}%
\usepackage{comment}
\usepackage{enumitem}
\usepackage{makecell}
\usepackage{fancybox}
\usepackage{placeins}
\usepackage{todonotes}
\usepackage{siunitx}
\usepackage{tabularx}
\usepackage{multirow}       
\usepackage{array}          
\usepackage{hyperref}
\usepackage{tikz}
\usepackage{subcaption} 
\usepackage{booktabs}
\usepackage{multirow}
\usepackage{siunitx}
\newcommand{\LLM}[1]{\multirow{3}{2.7cm}{\raggedright #1}}
\definecolor{pe}{RGB}{255,235,235}  
\newcommand{\sbelBoldNoIndent}[1]{\noindent{\textbf{#1}}}
\newcommand{\sbelItalicNoIndent}[1]{\noindent{\textit{#1}}}

\graphicspath{{../}}
\begin{document}

\title{ChronoLLM: Customizing Language Models for Physics-Based Simulation Code Generation}


\author{Jingquan Wang         \and
        Andrew Negrut \and 
        Harry Zhang \and
        Khailanii Slaton \and
        Shu Wang \and
        Radu Serban \and
        Jinlong Wu \and
        Dan Negrut
        }


\institute{
    Jingquan Wang \at
    Department of Mechanical Engineering, University of Wisconsin-Madison, 
    1513 University Avenue, 53706, Madison, USA \\
    \email{jwang2373@wisc.edu}
    \and 
    Andrew Negrut \at
    Department of Computer Science, Rice University, 
    6100 Main Street, 77005, Houston, TX, USA \\
    \email{agn5@rice.edu}
    \and 
    Khailanii Slaton \at
    Department of Mechanical Engineering, University of Wisconsin-Madison, 
    1513 University Avenue, 53706, Madison, USA \\
    \email{kslaton@wisc.edu}
    \and 
    Shu Wang \at
    Department of Mechanical Engineering, University of Wisconsin-Madison, 
    1513 University Avenue, 53706, Madison, USA \\
    \email{swang597@wisc.edu}
    \and 
    Radu Serban\at
    Department of Mechanical Engineering, University of Wisconsin-Madison, 
    1513 University Avenue, 53706, Madison, USA \\
    \email{serban@wisc.edu}
    \and 
    Jinlong Wu \at
    Department of Mechanical Engineering, University of Wisconsin-Madison, 
    1513 University Avenue, 53706, Madison, USA \\
    \email{jinlong.wu@wisc.edu}
    \and
    Dan Negrut \at
    Department of Mechanical Engineering, University of Wisconsin-Madison, 
    1513 University Avenue, 53706, Madison, USA \\
    \email{negrut@wisc.edu}
}

\date{Received: date / Accepted: date}

\maketitle

\begin{abstract}
This contribution is concerned with the following issue: can pretrained large language models (LLMs) be refined and customized to the point where they become virtual assistants helping experts with the effective use of a simulation tool? In this case study, the ``simulation tool'' considered is PyChrono, an open source multi-physics dynamics engine for multibody systems. We present a framework for refining and customizing both open- and closed-source LLMs to harness the power of AI in generating scripts that perform PyChrono virtual experiments. We refine and customize several classes of LLMs through a process that leads to a quantifiable improvement in the quality of the generated PyChrono simulation scripts. These scripts can range from simple single-pendulum simulations to complex virtual experiments involving full vehicles on deformable terrain. While the generated scripts are rarely perfect, they often serve as strong starting points for the user to modify and improve on. Additionally, the LLM can answer specific API questions about the simulator, or recommend modeling approaches. The framework discussed is general and can be applied to lower the entry barrier for simulation tools associated with other application domains.
\subclass{MSC code1 \and MSC code2 \and more}
\end{abstract}

\section{Introduction}
\label{sec:Introduction}
This contribution describes the process of \textit{refining} and \textit{customizing} a large language model (LLM) to improve its ability to generate PyChrono models \cite{pychrono2022}. Refining pertains to using fine-tuning, prompt engineering, and prompt learning, among other techniques. Customizing pertains to changing the LLM to be particularly good at generating PyChrono models. Overall, this effort is justified by the ultimate goal of having the LLM assist a Chrono \cite{chronoOverview2016} user in generating a digital twin of a mechanical system of interest, e.g., a three link robot arm, a rover possessing a collection of sensors, or a four-bar linkage mechanism. 

Since the word ``model'' is used in two contexts, their semantics are differentiated as follows. The PyChrono model that ChronoLLM produces represents a digital twin of a mechanical system. Carrying out a virtual experiment with this digital twin requires a PyChrono script, which includes the model and also specifies how the model is exercised in a virtual experiment that presumably mirrors a physical experiment. PyChrono is used to run the script that outputs the time evolution of the digital twin. Ideally, the time evolution of the digital twin should be close to that of its physical counterpart to result in a small sim-to-real gap. Conversely, the AI model that we seek to refine and customize is an LLM, which has billions to trillions of weights that the techniques discussed herein may or may not change.

The choice of focusing on LLMs that produce PyChrono digital twins is motivated by the fact that the authors are involved in various aspects of augmenting the Project Chrono simulation infrastructure. However, the framework to produce ChronoLLM can be adopted and adapted to work in conjunction with any other open or closed-source simulators from other domains.

\subsection{Brief Overview of Project Chrono}
\label{sec:chrono}
Project Chrono \cite{chronoOverview2016} is an open-source, physics-based simulation framework that supports the modeling, simulation, and analysis of mechatronic systems. It is designed for high-fidelity simulations, and released under a permissive BSD3 license that allows unfettered use, modification, and redistribution. PyChrono is the Python wrapper for Project Chrono, providing a user-friendly interface to the core functionalities of Project Chrono. It allows users to leverage the power of Project Chrono using Python, making it accessible to a broader range of users who prefer scripting in Python over C++. At the time of publishing, PyChrono has been installed more than 140000 times \cite{pychrono-condaforge}.

Project Chrono encompasses a wide range of features, and PyChrono inherits a subset of these capabilities:
\begin{enumerate}
    \item Chrono::Engine: Provides core functionality for multibody dynamics and nonlinear finite element analysis, with robust treatment of friction and contact using both the penalty method and the Lagrange-multiplier method.
    \item Chrono::FSI: Enables the simulation of fluid-structure interaction problems. 
    \item Chrono::Vehicle: Includes a comprehensive library of wheeled and tracked vehicles, facilitating high-fidelity vehicle dynamics simulations, engine modeling, terrain-tire interaction, and deformable terrain simulations. It focuses on off-road and unstructured scenarios involving deformable terrains and movable obstacles.
    \item Chrono::ROS: A bridge that enables Chrono to interact with the autonomy stack or sensor infrastructure of robots using ROS2. 
    \item Chrono::Sensor: Provides sensor modeling and simulation.
    \item Chrono::DEM: Enables the simulation of discrete element method problems, with a focus on granular materials.
    \item Chrono::CRM: supports terramechanics simulations enabling one to model the interaction between a deformable terrain and a robot, vehicle, or rover.
    \item Chrono::Parsers: A tool to import external models and to interact with other languages, e.g., URDF.
\end{enumerate}

Chrono has been utilized in conjunction several NASA project, e.g., the VIPER mission, the RASSOR excavator, the Moon Ranger rover, the Moon Racer astronaut vehicle, and the IM-4 lunar lander. It has been adopted by the Department of Defense High-Performance Computing Modernization Program (HPCMP) for the simulation of wheeled and tracked vehicles in the CREATE-GV project \cite{skorupaGVSETS2017}. Additionally, Chrono has been tested in NATO benchmarking exercises for off-road vehicle mobility analysis \cite{NATObenchmark2018}. Other notable users include the Jet Propulsion Lab, U.S. Army, National Renewable Energy Lab (NREL), National Higher French Institute of Aeronautics and Space, European Space Agency, Japan Aerospace Exploration Agency (JAXA), to name a few. Applications of Chrono span extraterrestrial exploration \cite{toso2015esa,ferrari2017n,jplREACH2020,regolithSamplingChrono2023}, machine learning in robotics \cite{cookNeuralNets2017}, image processing \cite{mccormacChrono2017,xu2018ComputerVisionChrono}, autonomous vehicles \cite{batteryAutonomousVehiclesChrono2016,goodin2017unmanned,haraus2017performance,trajectoryPlanningEpureanu2023,tractionControlChronoHungary2024,roboticFishChrono2024}, tracked-vehicle design \cite{jonakTrackedVehicle2018}, fluid-solid interaction \cite{dualSPHCouplingChrono2016}, bridge suspension \cite{wangCableChrono2019}, hardware-in-the-loop simulations \cite{HILchrono2019}, wind turbine dynamics \cite{perezChrono2018,marten2019benchmark}, hydrodynamics \cite{ogden2023hydrochrono} and oil industry applications \cite{oilIndustryChrono2020}.

As of Jun 2025, Project Chrono has been developed for more than 25 years and iterated to its 9.0.0 version. It has over 750 forum members \cite{projectChronoForum}, the project has been starred with more than 2400 times, and forked more than 400 times on GitHub \cite{projectChronoGithub}, resulting in numerous derivative projects. The Chrono GitHub repository is ranked amongst the first 0.014\% of the repositories on GitHub (rank 22,774 out of 158,541,903 public GitHub repositories worldwide. 

Chrono is developed jointly at the University of Wisconsin--Madison (USA) and the University of Parma (Italy). Because it is ($i$) an academic open-source code at the forefront of research in digital twin technology, ($ii$) developed primarily within university settings, and ($iii$) freely available --- creating a sophisticated graphical user interface and comprehensive documentation has not been a top priority owing to lack of resources. As a result, Chrono can be challenging to learn and adopt. Despite having more than 3500 webpages detailing the simulator's use and features, the Chrono documentation in several areas remains sparse. Improving the user experience is a long-term goal of the Chrono development team. This contribution reports on our efforts to leverage LLMs to assist users in setting up PyChrono virtual experiments more effectively.

\subsection{A Brief Overview of LLMs and Domain-Specific LLMs}
\label{sec:domainLLMs}
Recent advancements in artificial intelligence (AI) have led to breakthroughs in natural language processing. As highlighted in studies that investigate how LLM performance scales with model size \cite{kaplan2020scalinglaw1,hoffmann2022scalinglaw2}, as their size increases, LLMs exhibit emergent abilities that were not evident at smaller scales. These emergent abilities include enhanced comprehension, reasoning, and language generation \cite{wei2022emergent1,schaeffer2024emergent2}, paving the way for LLMs to extend their utility beyond simple language tasks to more complex applications across various domains.

In science and engineering, LLMs are transforming how experts approach problem-solving and design. Advanced closed-source LLMs, such as the GPT family \cite{OPENAI2024gpt4,NIPS2020} by OpenAI, the Gemini family \cite{team2023gemini} by Google, and the Claude family \cite{Claudemodelcard} by Anthropic, have been demonstrating steady progress in handling complex tasks like code generation. The usage of these closed-source LLMs is typically fee-based, and is done via websites and/or API calls. While these LLMs, trained on trillions of tokens from publicly available data, perform exceptionally well on general tasks, they have limitations when deployed in domain-specific applications, e.g., the task of generating a model that can be run in Chrono. This is caused by two main reasons: due to their large size, frequent retraining to incorporate recent developments is impractical; and,  lack of exposure to up-to-date information can lead to inaccurate responses, particularly in domain-specific information processing where LLMs may not fully understand new terminology. To counteract this and build an LLM that can help with domain-specific problems, it is recommended to customize the LLM using specialized knowledge, enabling it to adapt and accurately comprehend new (unseen) information. 

\begin{table}[!t]
\centering
\begin{tabular}{|l|m{4.5cm}|m{3.5cm}|m{3.0cm}|}
	\hline
	& \textbf{Prompt Engineering} & \textbf{Prompt Learning} & \textbf{Fine-Tuning} \\ \hline
	\textbf{Tech} & 
	\begin{itemize}[leftmargin=0.5cm]\vspace{3pt}
	{\footnotesize
	\item Few-shot learning \cite{wang2020fslearning} 
	\item Chain-of-thought reasoning \cite{wei2022NIPS_ChainOfThought} \vspace{-3pt}
	\item Scratchpad reasoning \cite{google2021scrathpad} \vspace{-5pt}
	}
	\end{itemize} & 
	\begin{itemize}[leftmargin=0.5cm]\vspace{3pt}
	{\footnotesize
	\item Prompt tuning \cite{gu2021ppt}
	\item P-tuning \cite{liu2021ptun} \vspace{-3pt}
	\item Prefix tuning \cite{li2021prefix} \vspace{-5pt}
	}
	\end{itemize} & 
	\begin{itemize}[leftmargin=0.5cm]\vspace{3pt}
	{\footnotesize
	\item Adapters \cite{hu2023adapter}
	\item LoRA \cite{hu2022lora} \vspace{-3pt}
	\item RLHF \cite{bai2022rlhf} \vspace{-5pt}
	\item Supervised fine-tuning \cite{dodge2020fine} \vspace{-3pt}
	}
	\end{itemize} 
	\\ \hline \vspace{3pt}
	\textbf{Pros} & 
	\begin{itemize}[leftmargin=0.5cm] \vspace{3pt}
	{\footnotesize
	\item Lowest investment 
	\item Least expertise \vspace{-3pt}
	\item Retains old skills \vspace{-7pt}
	}
	\end{itemize} & 
	\begin{itemize}[leftmargin=0.5cm] \vspace{3pt}
	{\footnotesize
	\item Lower investment 
	\item Better performance \vspace{-3pt}
	\item Retains old skills \vspace{-7pt}
	}
	\end{itemize} & 
	\begin{itemize}[leftmargin=0.5cm]
	{\footnotesize
	\item Best performance \vspace{-7pt}
	}
	\end{itemize}
	\\ \hline 
	\textbf{Cons} & 
	\begin{itemize}[leftmargin=0.5cm]
	{\footnotesize
	\item Worst performance
	\item Slow down inference \vspace{-3pt}
	\item Limited addition of skills \vspace{-7pt}
	}
	\end{itemize} & 
	\begin{itemize}[leftmargin=0.5cm]
	{\footnotesize
	\item Slow down inference \vspace{-3pt}
	\item Limited addition of skills \vspace{-7pt}
	}
	\end{itemize} & 
	\begin{itemize}[leftmargin=0.5cm]\vspace{3pt}
	{\footnotesize
	\item Medium investment 
	\item Longer training 
	\item More expertise \vspace{-3pt}
	\item Forgetting old skills \vspace{-7pt}
	}
	\end{itemize}  \\ \hline
\end{tabular}
\caption{Comparison of Methodologies to customized LLMs from pretrained LLMs}
\label{tab:fine_tune_methods}
\end{table}

From a high vantage point, there are four ways to produce a domain-specific LLM. 

\begin{enumerate}
\item Training from Scratch: This approach involves collecting extensive domain-specific data and training a new LLM tailored to address specific problems. For instance, in the field of high-performance computing, models such as OMPGPT-0.78B \cite{chen2024ompgpt} and MonoCoder-0.89B \cite{kadosh2023domain} have claimed superior performance compared to GPT-3.5 in parallelizing C++ code using OpenMP by training on numerous examples of OpenMP programs. The primary advantage of this method is the potential for achieving the highest performance when (a) sufficient data, (b) computational resources, and (c) expertise are available. However, the process is often daunting, including the high cost of data collection, model training, and hardware procurement. Training even a small-scale 7-8B LLM like LLaMA3-8B \cite{llama3modelcard} can cost millions of dollars \cite{trainingCosts-2025}. If the domain-specific LLM is too small, its performance on specific tasks will be highly constrained.

\item Prompt engineering, including in-context learning (few-shot prompting with domain-specific examples): this involves structuring model inputs to guide reasoning. Techniques such as chain-of-thought (CoT) prompting \cite{wei2022NIPS_ChainOfThought} and ScratchPad \cite{google2021scrathpad} encourage LLMs to generate intermediate reasoning steps explicitly, improving problem-solving and reasoning abilities in complex tasks. In the multibody dynamics field, researchers have used this approach in conjunction with GPT-4 to generate models for the Exudyn library \cite{gerstmayr2023exudyn,gerstmayr2024multibodyllm}. Other notable examples include MyCrunchGPT \cite{kumar2023mycrunchgpt} for generating Python codes for physics-informed neural networks (PINN) \cite{RAISSI2019PINN}, and GeoGPT \cite{KIM2024geogpt} for generating Matlab code in geotechnical engineering. The advantages of this approach are that (a) minimal expertise and computational resources are required, (b) there is virtually no computation spent on altering the model parameters, and (c) GPT-4 is a versatile model to start with. Some of the limitations include potential security and privacy issues due to the need to upload information to the LLM server, limited customization capabilities, and high inference cost and  latency.

\item Prompt Learning: these are approaches that extend traditional prompt engineering by incorporating trainable parameters to optimize prompt representations, improving model adaptability. Methods such as prefix-tuning \cite{li2021prefix}, P-tuning \cite{liu2021ptun}, and prompt tuning \cite{gu2021ppt} adjust learned prompt embeddings (trainable prompt representations in a neural network, stored as numerical vectors instead of human-readable text) to better steer the model's behavior while keeping the core model parameters frozen. For example, prompt learning has been used to customize LLMs for clinical diagnosis \cite{taylor2023clinical}. The pros of this approach are its low training cost and the fact that the LLM parameters remain unchanged, often yielding better results than simple prompt engineering. The cons include the limited ability to improve performance without altering all model parameters and a slight inference slowdown.

\item Fine-Tuning and Parameter-Efficient Fine-Tuning: Looking beyond closed-source LLMs, the availability of open-source LLMs with accessible model weights, such as those in the LLaMA \cite{meta2023llama,touvron2023llama2,roziere2023codellama,llama3modelcard}, Mistral \cite{jiang2023mistral}, Gemma \cite{google2024gemma,codegemma_2024}, and Phi \cite{abdin2024phi3} families, opens fine-tuning opportunities. Fine-tuning involves starting from pretrained LLM weights and continuing training (changing the model weights) using domain-specific data. As demonstrated by various studies, fine-tuning works really well for customizing LLMs for domain problems. For example, OceanGPT \cite{bi2023oceangpt} is trained based on LLaMA3-8B for ocean-related problems, and Hippocrates \cite{acikgoz2024healthcarellm} fine-tuned Mistral-7B and LLaMA2-7B, showing superior performance compared to larger models up to 70B. AnomalyGPT \cite{gu2023anomalyagpt} fine-tuned large vision-language models for anomaly detection in manufacturing processes. C4QGPT, supporting query generation for quantum dynamics \cite{aragones2024c4qgpt}, is fine-tuned based on BERT \cite{bertGoogle2018}, and MPIrigen \cite{schneider2024mpirigen} fine-tuned MonoCoder to parallelize C++ code using MPI. The advantages of fine-tuning include achieving the best results based on pretrained LLMs, the ability to incorporate extensive data, and not slowing down inference. However, fine-tuning requires non-trivial amounts of data, necessitates expertise, and may lead to forgetting old skills or even catastrophic forgetting. In \cite{zhao2024loraland}, the authors showed that parameter-efficient fine-tuning (PEFT) with open-source 7-8B LLMs can outperform GPT-4 on domain-specific tasks by an average of 10 points. Several studies showed that the performance improvements obtained by fine-tuning exceed those gained through prompt engineering \cite{liu2022finetunevsicl,mosbach2023finetunevsicl2}. 
\end{enumerate}

The advantages and disadvantages of different customization methods with a pretrained LLM in are summarized in Table \ref{tab:fine_tune_methods}. The domain-specific LLMs mentioned above are summarized in Table \ref{tab:engineering_llms}.

\begin{table}[ht]
\centering
\begin{tabular}{|c|c|c|c|c|}
\hline
\textbf{LLMs} & \textbf{Size} & \textbf{Base Model} & \textbf{Method} & \textbf{Domain} \\
\hline
OPENMPGPT\cite{chen2024ompgpt} & 0.78B & N/A & TS & HPC \\
\hline
MonoCoder\cite{kadosh2023domain} & 0.89B & N/A & TS & HPC \\
\hline
MyCrunchGPT\cite{kumar2023mycrunchgpt} & N/A & GPT-4 & PE & PINN \\
\hline
Exudyn \cite{gerstmayr2024multibodyllm} & N/A & GPT-4 & PE & Multibody Dynamics \\
\hline
GeoGPT\cite{KIM2024geogpt} & N/A & GPT-4 & PE & Geotechnical Engineering \\
\hline
Clinical Prompt learning\cite{taylor2023clinical} & N/A & N/A & PL & Health Care\\
\hline
C4QGPT\cite{aragones2024c4qgpt} & 2x0.11B & Bert\cite{bertGoogle2018} & FT & Quantum Dynamics \\

\hline
MPIRIGEN\cite{schneider2024mpirigen} & 0.89B & MonoCoder & FT & HPC \\
\hline
OceanGPT\cite{bi2023oceangpt} & 8B & LLaMA3-8B & FT & Ocean Engineering \\
\hline
Hippocrates\cite{acikgoz2024healthcarellm} & 7B & LLaMA2-7B & FT & Health Care \\
\hline
\end{tabular}
\caption{Overview of LLMs for domain-specific problems. Abbreviations: Trained from Scratch -- TS; Prompt Engineering -- PE; Prompt Learning -- PL; Fine-tuning -- FT; High Performance Computing -- HPC}
\label{tab:engineering_llms}
\end{table}

\subsection{PyChrono Challenges and current LLMs}
\begin{figure}[!t]
    \centering
    \includegraphics[width=12cm]{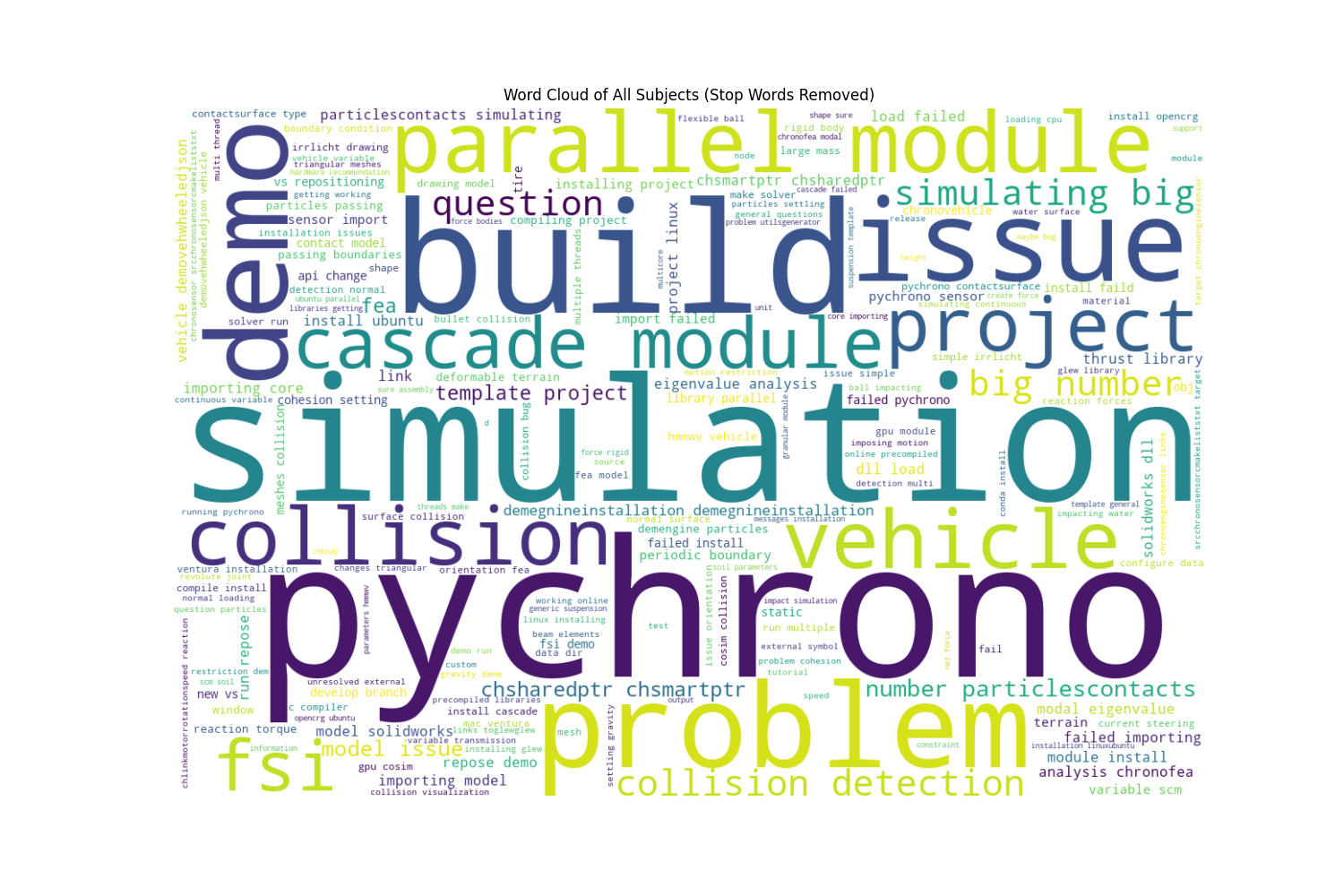}
    \caption{The word cloud of Project Chrono Forum}
    \label{fig:word_cloud}
\end{figure}

The goal of this effort is to develop a domain-specific LLM that increases user productivity by generating candidate PyChrono models in response to prompts describing the mechatronic system of interest. To that end, it is instructive to understand the issues the users struggle with. We conducted an analysis of the discussions in the Project Chrono user forum \cite{projectChronoForum}, which currently has more than 750 members. Up to May 2024, there have been 2576 multi-round conversations in the forum. Figure \ref{fig:word_cloud} shows a word cloud representing the most frequently mentioned words. Using Latent Dirichlet Allocation (LDA) \cite{sievert2014lda} to analyze the subject lines of these conversations, we identified three primary topics:

\begin{enumerate}
    \item Case-by-Case Simulation Setup Challenges: The most common topic, accounting for $38\%$ of the discussions, pertains to specific challenges encountered when setting up simulations with PyChrono. When users set up PyChrono models, they engage in discussions about identifying correct model parameter choices, handling unexpected simulation behaviors and crashes, and enhancing simulation performance. The LDA analysis suggests that common challenges involve collision detection, configuring the dynamics of physical interactions, and prescribing accurate motion to system bodies.
    
    \item API Usage Problems: Representing about $37\%$ of the forum discussions, this category reflects the difficulties users face with the extensive array of API elements provided by PyChrono. The size of the code and the complexity inherent in a multi-physics simulator has led to a number of 18320 API related assets (including functions, variables, typedefs, enumerations) as per the latest version of Project Chrono \cite{chronoAPIWebSite}. This extensive API poses usability challenges, particularly for newcomers or those engaged in complex simulation tasks. Users often report difficulties in identifying the right API calls or employing them effectively within their projects. Additionally, the LDA findings suggest that these issues are compounded when addressing bugs, especially in simulations involving terrain and vehicles. Commonly reported API-related challenges include difficulties with tire models, vehicle modeling, collision detection, and boundary conditions. Compounding the problem is the fact that the API has changed repeatedly over time, increasing the likelihood that LLMs lack exposure to the most recent updates.
    
    \item Module Installation and Compilation Issues: Making up approximately $20\%$ of the conversations, this topic addresses the issues related to the installation and compilation of various PyChrono modules. Incorrect installation procedures and compatibility issues frequently pose obstacles for users, limiting their ability to effectively utilize PyChrono. The LDA analysis draws attention to the fact that these discussions frequently center on specific modules, notably those involving the PyChrono sensor module and CUDA dependencies. Users often run into issues arising from cross-operating system compatibility, e.g., Windows vs. Linux vs. MacOS.
 
\end{enumerate}

\paragraph{The limitations of current LLMs for Chrono-related problems}
Despite recent technological advances, modern LLMs, when used ``out of the box,'' struggle to effectively address Chrono-related questions, often adding complexity rather than providing practical solutions. 
While some LLMs exhibit a basic understanding of PyChrono, see Table \ref{tab:LLM_Chrono_Knowledge} for context, their capabilities remain inadequate for generating reliable simulation code or handling complex tasks. 
Specifically, out of the box, LLMs have only a basic understanding of the Chrono API. This is largely due to the frequent changes in the API (see, for example, Table \ref{tab:api_changes}), which disrupt the pretraining process. Beyond this, the models are not supervised fine-tuned for Project Chrono, resulting in limited adaptation to the specific requirements of Chrono-related tasks. As a result, LLMs often fail to reflect recent API updates and may hallucinate method names or misuse interfaces for visualization, solver configuration, or collision handling.

\begin{table}[!b]
	\centering
	\caption{Baseline Chrono Knowledge in Various LLMs}
	\label{tab:LLM_Chrono_Knowledge}
	{\footnotesize
	\begin{tabular}{p{2cm} p{1.15cm} p{1.25cm} p{1.5cm} p{1.25cm} p{1.25cm} p{2cm}}
		\toprule
		\textbf{Model} & \textbf{GPT-4} & \textbf{GPT-3.5} & \textbf{LLaMA3-8B} & \textbf{Phi3-mini} & \textbf{Mistral-7B} & \textbf{CodeGemma-7B} \\
		\midrule
		Chrono Intro & $\checkmark$ & $\checkmark$ & $\checkmark$ & $\checkmark$ & $\checkmark$ & $\checkmark$ \\
		Chrono API & $\checkmark$ & $\checkmark$ & $\times$ & $\checkmark$ & $\checkmark$ & $\checkmark$ \\
		\bottomrule
	\end{tabular}
	}	
\end{table}

\begin{table}[h!]
\centering
\begin{tabular}{|l|l|}
\hline
\textbf{Old API} & \textbf{New API} \\ \hline

\multicolumn{2}{|c|}{\textbf{Basic Usage}} \\ \hline
\texttt{chrono.ChVectorD(...)} & \texttt{chrono.ChVector3d(...)} \\ \hline
\texttt{chrono.ChQuaternionD(...)} & \texttt{chrono.ChQuaterniond(...)} \\ \hline
\texttt{chrono.ChCoordsysD(...)} & \texttt{chrono.ChCoordsysd(...)} \\ \hline
\texttt{sys.Set\_G\_acc(...)} & \texttt{sys.SetGravitationalAcceleration(...)} \\ \hline
\texttt{box1.SetPos\_dt(...)} & \texttt{box1.SetPosDt(...)} \\ \hline
\multicolumn{2}{|c|}{\textbf{Collision and Contact System}} \\ \hline
\texttt{chrono.ChCollisionSystemBullet(...)} & \texttt{sys.SetCollisionSystemType(...)} \\ \hline
\texttt{ground.SetBodyFixed(...)} & \texttt{ground.SetFixed(...)} \\ \hline
\texttt{ground.SetCollide(...)} & \texttt{ground.EnableCollision(...)} \\ \hline
\texttt{sys.GetNcontacts(...)} & \texttt{sys.GetNumContacts(...)} \\ \hline
\texttt{chrono.ChMaterialSurfaceNSC(...)} & \texttt{chrono.ChContactMaterialNSC(...)} \\ \hline
\texttt{chrono.ChMaterialSurfaceSMC(...)} & \texttt{chrono.ChContactMaterialSMC(...)} \\ \hline
\texttt{chrono.CastToChMaterialCompositeNSC(...)} & \texttt{chrono.CastToChContactMaterialCompositeNSC(...)} \\ \hline
\multicolumn{2}{|c|}{\textbf{Solver Settings}} \\ \hline
\texttt{sys.SetSolverMaxIterations(...)} & \texttt{sys.GetSolver().AsIterative().SetMaxIterations(...)} \\ \hline
\texttt{sys.SetSolverForceTolerance(...)} & \texttt{sys.GetSolver().AsIterative().SetTolerance(...)} \\ \hline
\multicolumn{2}{|c|}{\textbf{Visualization}} \\ \hline
\texttt{chrono.ChFrameD(...)}&\texttt{chrono.ChFramed(...)}\\ \hline
\texttt{pend\_2.SetFrame\_COG\_to\_REF(...)} & \texttt{pend\_2.SetFrameCOMToRef(...)} \\ \hline
\texttt{pend\_2.SetFrame\_REF\_to\_abs(chrono.ChFrameD(...))} & \texttt{pend\_2.SetFrameRefToAbs(chrono.ChFramed(...))} \\ \hline
\texttt{chrono.ChCylinderShape(...)} & \texttt{chrono.ChVisualShapeCylinder(...)} \\ \hline
\texttt{ground.AddVisualShape(...)} & \texttt{ground.AddVisualShape(..., chrono.ChFramed(...))} \\ \hline
\texttt{veh.ChIrrGuiDriver(...)} & \texttt{veh.ChInteractiveDriverIRR(...)} \\ \hline

\end{tabular}
\caption{Part of the API changes from old to new versions in PyChrono}
\label{tab:api_changes}
\end{table}

 Addressing the limitations pointed out above requires more than prompt engineering or prompt learning techniques. Considering the complexity of PyChrono and the need for accurate, domain-specific outputs, we opted to fine-tune existing LLMs, particularly focusing on their adaptation to the unique aspects of the Chrono tasks. As demonstrated in this work, fine-tuning yields measurable performance gains over approaches based solely on prompt engineering or prompt learning.

\section{Methodology}
\label{sec:Methodology}
\subsection{Problem Statement}
\begin{figure}
    \centering
    \includegraphics[width=15cm]{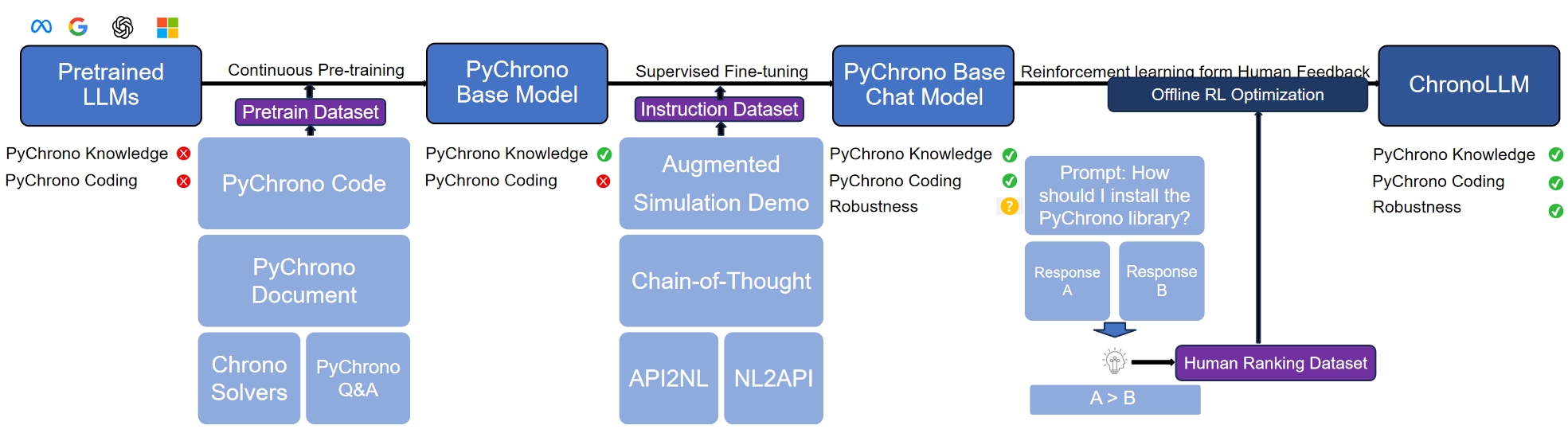}
    \caption{The whole pipeline of ChronoLLM to customize open-source LLMs for PyChrono tasks.}
    \label{fig:pipline}
\end{figure}

The ChronoLLM pipeline is shown in Fig \ref{fig:pipline}. The training of PyChrono knowledge and coding ability consists of two main stages: continued pretraining and instruction fine-tuning. 
For the reinforcement from human feedback (RLHF) component, it is important to note that we do not employ online RLHF approaches such as PPO \cite{NIPS2022HumanFeedback} or GRPO \cite{deepseek2024}, which are typically aimed at improving coding ability. 
Instead, we adopt offline RL methods, in particular SimPO \cite{meng2024simpo}, as they are better suited for robust Q\&A tasks, such as providing PyChrono installation guidance.
From a high-vantage point, we start with a pretrained ``base model'' and then improve it. The starting LLM can be one of many available pretrained models, regardless of whether it is open-source or closed-source. If the model is open-source and of reasonable size, we can train locally. Otherwise, the training process will require a financial investment since (a) the model is trained in the cloud using hardware assets that require pay-as-you-go fees, and/or (b) there are costs associated with API calls for a non-open source model. While the figure presents the model under the generic label ``ChronoLLM,'' its proper designation depends on the underlying pretrained model -- for instance, ChronoLLaMA3-70B, ChronoChatGPT-4, or ChronoGemma2-27B.

\subsection{Selection of Base Models}  
LLMs can be broadly categorized into two architectural types: 
\emph{dense models}, where all parameters are activated during inference, and \emph{sparse models}, such as Mixture of Experts (MoE), where only a subset of parameters is used per input. 
Representative MoE models include the Mixtral family \cite{jiang2024mixtral} by Mistral AI and DeepSeek-MoE \cite{deepseek2024}, 
while examples of dense models include LLaMA3-70B, GPT-4o, and Gemma2-27B. Although there is speculation that GPT-4 employs an MoE architecture, its exact design has not been publicly disclosed.

Since MoE models activate only a subset of their parameters (experts) for each input, this reduces computational costs during inference, and allows for better scaling, i.e., assembling larger models, without a proportional increase in computational requirements. While MoE models often offer faster inference at comparable sizes, they tend to lag behind dense models in accuracy. Moreover, training and managing MoE models can be complex. Dense models, in contrast, are easier to train and provide more robust, uniform performance, albeit at a higher computational cost during inference.

The consistent accuracy, reliability, and robust performance of dense models make them the ideal choice for fine-tuning on PyChrono-related tasks. The simplicity of training dense models and their superior generalization across tasks further support their selection over MoE-based alternatives. Considering these trade-offs, we selected dense models. The chosen dense models were evaluated based on their performance in widely recognized benchmarks, such as the HumanEval benchmark for coding tasks \cite{chen2021evaluating1} and the MMLU benchmark for general reasoning tasks \cite{hendrycks2020mmlu}. Specifically, in this contribution we adopt the following dense LLMs:
\begin{itemize}
    \item \textbf{GPT-4o} and its lightweight variant \textbf{GPT-4o mini}, for their good coding performance and general good knowledge about Project Chrono,
    \item \textbf{LLama3.1-8B}, \textbf{LLaMA3.3-70B} and \textbf{Llama-4-Scout-17B-16E}, which are open-sourced models from 8B to 109B that offer good performance on diverse tasks,
\end{itemize}
The decision to exclude certain dense models, such as CodeLLaMA \cite{roziere2023codellama} and StarCoder \cite{li2023starcoder_LLM}, is based on their limitations in our context. While CodeLLaMA is specialized for coding tasks, it is built on the older LLaMA2 architecture and underperforms relative to LLaMA3-70B on the HumanEval benchmark \cite{humanEvalBench2021}. Similarly, StarCoder models, despite their multi-language coding specialization, do not align with our use case, as Chrono users do not require a tool exclusively focused on code generation.

\subsection{Continual Pretraining}
The pretraining stage of LLMs is predominantly unsupervised, allowing models to process vast datasets and learn underlying patterns, structures, and knowledge. During this stage, the model acquires foundational knowledge that underpins its capabilities across a wide range of tasks. Pretraining typically involves either causal language modeling (CLM) or masked language modeling (MLM), depending on the architecture. In CLM, the model is trained to predict the next token in a sequence based on preceding tokens. In MLM, a subset of tokens in the input is randomly masked, and the model learns to predict the missing tokens based on surrounding context.

In our approach, we continue pretraining the model using the standard CLM approach. Given an input token sequence $\bm{x} = (x_0, x_1, x_2, \ldots)$, the model is trained to predict each token $x_i$ in an autoregressive manner. The objective is to minimize the negative log-likelihood, defined as:
\begin{equation}\label{eq_clm}
    \mathcal{L}_{\textrm{CLM}} (\Theta) = \mathbb{E}_{\bm{x} \sim \mathcal{D}_{\textrm{PT}}}\left[ -\sum_i \log p(x_i \mid x_0, x_1, \ldots, x_{i-1}; \Theta)\right] \; .
\end{equation}
Here, $\Theta$ represents the model parameters, $\mathcal{D}_{\textrm{PT}}$ is the pretraining dataset, $x_i$ is the token to be predicted, $x_0, x_1, \ldots, x_{i-1}$ form the context, and $p(x_i \mid x_0, x_1, \ldots, x_{i-1}; \Theta)$ is the probability of getting token $x_i$ given the context $x_0, x_1, \ldots, x_{i-1}$ under a model parameterized by $\Theta$.

The LLMs acquire knowledge during pretraining, but certain emergent abilities -- advanced comprehension, reasoning, and language generation -- become apparent only when the pretraining loss falls below a critical threshold \cite{du2024understandemergent}. These emergent abilities enhance the model's capacity to follow instructions during fine-tuning. To ensure that the model gains specialized knowledge relevant to Chrono, we employ continual pretraining on Chrono-related materials, specifically focusing on the PyChrono API documentation, example scripts, and user queries.

Continual pretraining involves further training a pretrained model on new, domain-specific data. This approach presents challenges, particularly the risk of catastrophic forgetting \cite{french1999catastrophic}, where the model loses previously acquired knowledge while adapting to account for new data. To mitigate this risk, several strategies can be employed \cite{wu2022pretrained}:
\begin{itemize}
    \item \textbf{Warm-up Strategies}: As recommended in \cite{gupta2023continualwarm}, gradually increase the learning rate as one picks up and continues the pretraining, a step that helps the model adapt to new data without overwhelmingly lose prior knowledge.
    \item \textbf{Regularization Techniques}: Methods like elastic weight consolidation (EWC) penalize changes to critical weights, preserving previously learned information \cite{kirkpatrick2017overcoming}.
    \item \textbf{Replay Methods}: Mixing new data with previously seen data during training reinforces older knowledge while integrating new information \cite{lopez2017gradient}.
\end{itemize}
We deemed regularization techniques as too sophisticated and cumbersome -- it is not clear how this technique can be employed for closed-source LLMs. In the end, for our application, only the warm-up strategy is feasible because, for most open-source LLMs, only the model weights are available, while the original training data is not. By employing the warm-up strategy, we can continue pretraining LLMs to provide them with specialized Chrono-related knowledge while ensuring the retention of their general capabilities. This approach ensures the model becomes more proficient in domain-specific tasks without compromising its broad utility.

\subsubsection{Continual Pretraining Dataset}

We undertook a curation process to create a comprehensive pretraining dataset, which eventually included:
\begin{enumerate}
    \item \textbf{PyChrono Code Examples}: this called for cleaning and extracting representative code samples from various PyChrono projects using scripts. Preferably, the dataset covers a wide range of use cases captured in diverse and high-quality learning material. A ``high quality''  example is one that has good comments and follows a set of good practices associated with the task of generating a PyChrono simulation scenario via a Python file.

    \item \textbf{PyChrono Documentation}:  
    Include foundational documents, such as installation guides, introductory materials, and basic tutorials. The challenge is to have a consistent set of documents that reference the most recent API and simulation-setup good practices. The Chrono body of documentation spans more than 3500 webpages.

    \item \textbf{PyChrono Q\&A}:  
    Compile questions and answers from PyChrono user forums and support channels. This data exposes the model to common issues, troubleshooting methods, and practical solutions. In this regard, the current collection is relatively limited.

    \item \textbf{Chrono Solvers}:  
    Although not immediately essential, materials related to Chrono solvers -- such as solver documentation, setting numerical integration parameters, pros and cons of variou frictional contact models, etc. -- can be included to enhance the LLM's understanding of simulation and solver techniques. These materials are valuable for improving the LLM's ability to handle complex simulation tasks. The current collection is relatively limited.
\end{enumerate}
By integrating these components, the dataset ensures the model receives a comprehensive and in-depth education in PyChrono. Table~\ref{tab:pretrain_samples} provides examples from each category, formatted in JSON \cite{json} for compatibility with existing LLM training pipelines.

\begin{table}[h!]
\centering
\renewcommand{\arraystretch}{1.5} 
\begin{tabular}{|p{4cm}|p{10cm}|}
\hline
\textbf{Category} & \textbf{Sample} \\ \hline

\textbf{PyChrono Code Examples} & 
\texttt{
\{ "text": "import pychrono as chrono  import pychrono.fea as fea  import pychrono.pardisomkl as mkl..." \}
} \\ \hline

\textbf{PyChrono Documentation} & 
\texttt{
\{ "text": "Overview of vehicle modeling and simulation The Chrono::Vehicle module provides templates for various..." \}
} \\ \hline

\textbf{PyChrono Q\&A} & 
\texttt{
\{ "text": "Is Chrono free? The entire Chrono software infrastructure is open source and released under a BSD-3 license. ..." \}
} \\ \hline

\textbf{Chrono Solvers} & 
\texttt{
\{ "text": "// PROJECT CHRONO - http://projectchrono.org // Chrono solvers based on Eigen iterative linear solvers..." \}
} \\ \hline

\end{tabular}
\caption{Examples of JSON-formatted samples for continual pretraining in different categories.}
\label{tab:pretrain_samples}
\end{table}

\subsection{In-Context Learning}
\label{subsec:in_context_learning}
In addition to refining and customizing LLMs through continued pretraining and fine-tuning -- the two approaches emphasized in this contribution -- we also consider in-context learning as a comparative baseline in our evaluation. In-context learning is a non-parametric strategy wherein a pretrained LLM is prompted with carefully selected examples and documentation, without altering its internal weights. While this method does not support persistent adaptation, it is appealing due to its low cost, ease of use, and ability to leverage powerful commercial LLMs out-of-the-box. In our case, the LLM is provided with PyChrono-specific prompts, including API documentation and code examples, to assess its ability to generate functional simulation scripts without explicit retraining. This strategy serves as a lightweight alternative to the more computationally intensive fine-tuning approach discussed in the following section.

\section{Fine-Tuning Methodology}
Unlike pretraining, which focuses on learning general patterns from large datasets, fine-tuning adjusts a pretrained model's parameters to optimize its performance for a specific task or domain, typically requiring less data. Fine-tuning is often the most critical step in adapting LLMs for domain-specific applications. While Chain-of-Thought (CoT) reasoning and its structured variants \cite{li2023structuredCOT} are prompting techniques rather than fine-tuning methods, they have been shown to enhance LLM generation abilities. To leverage this, fine-tuning can incorporate CoT-style reasoning, a process known as CoT-enhanced fine-tuning, where models are explicitly trained on reasoning-based demonstrations. This approach refines the model's ability to generate structured reasoning steps and improves task execution capabilities by aligning model outputs with desired responses.

\subsection{Supervised Fine-Tuning (SFT)}

Supervised Fine-Tuning (SFT) entails modifying the entire set of model parameters to improve task-specific performance. Although it often yields strong results, it is computationally intensive. The objective of SFT is to minimize the loss calculated on the output portion of the training sequence:
\begin{align}
    \mathcal{L}_{\textrm{SFT}}(\Theta) = \mathbb{E}_{\bm{x} \sim \mathcal{D}_{\textrm{SFT}}} \left[ -\sum_{i \in \textit{\{output\}}} \log p(x_i \mid x_0, x_1, \ldots, x_{i-1}; \Theta) \right] \; ,
\end{align}
where $\Theta$ denotes the model parameters, $\mathcal{D}_{\textrm{SFT}}$ represents the fine-tuning dataset, and $\bm{x} = (x_0, x_1, \ldots)$ refers to the tokenized input sequence.

\subsection{Parameter-Efficient Fine-Tuning (PEFT)}
Since SFT updates all model parameters, the computational cost can be prohibitive, particularly for large-scale LLMs. Furthermore, domain-specific data is often limited compared to the vast general-purpose datasets used in pretraining, making full fine-tuning impractical in many cases. Parameter-efficient fine-tuning (PEFT) offers a more resource-efficient alternative by modifying only a subset of parameters while preserving most of the pretrained model's knowledge. PEFT methods either update a subset of the model's parameters or introduce lightweight external modules to enhance efficiency \cite{he2021towards}. These techniques have been widely adopted in LLM applications and include:
\begin{itemize}
    \item \textbf{Adapters}: Trainable modules inserted into the model's architecture \cite{hu2023adapter, zhang2023adapter2}.
    \item \textbf{Soft Prompts}: Task-specific embeddings added to the input sequence \cite{lester2021softprompt}.
    \item \textbf{Selective Updates}: Methods like BitFit \cite{zaken2022bitfit}, which update only specific parameters, such as bias terms, while freezing the rest of the model.
\end{itemize}
Among PEFT techniques, Low-Rank Adaptation (LoRa) \cite{hu2022lora} stands out as a particularly effective method.

\subsubsection{LoRa and Its Variants}
LoRa employs low-rank decomposition to parameterize weight updates. For a pretrained weight matrix $W_0 \in \mathbb{R}^{d \times k}$, the weight update is defined as:
\begin{equation}
W_0 + \Delta W = W_0 + BA \; ,
\end{equation}
where $B \in \mathbb{R}^{d \times r}$ and $A \in \mathbb{R}^{r \times k}$, and $r \ll \min(d, k)$. During training, $W_0$ remains frozen, and only $A$ and $B$ are trainable. The modified forward pass for an input vector $x$ is
\begin{equation}
h = W_0 x + \Delta W x = W_0 x + BA x \; .
\label{eq:lora}
\end{equation}
LoRa is ideal for tasks requiring subtle adaptations to the attention mechanism. Variants such as LoRA+ \cite{hayou2024lora+}, QLoRA \cite{dettmers2024qlora}, and GaLoRa \cite{zhao2024galore} extend its capabilities through differential learning rates, quantization, and global attention adjustments, respectively.

\subsubsection{Advantages of PEFT Methods}
The benefits associated with PEFT techniques are:
\begin{itemize}
    \item \textbf{Storage Efficiency}: Adds only a small number of parameters, reducing storage requirements.
    \item \textbf{Memory Efficiency}: Requires less memory, enabling training on resource-constrained devices.
    \item \textbf{Computational Efficiency}: Low-rank updates reduce overhead, allowing faster training and inference.
\end{itemize}

\subsection{Dataset Preparation for Fine-Tuning}
The fine-tuning dataset was assembled to capture high-quality examples of PyChrono simulations that reflect best practices in model setup and usage. 

\medskip

\sbelBoldNoIndent{Data Collection and Generation.} Data is sourced from a combination of automated and manual methods:
\begin{itemize}[nosep]
    \item \textit{Descriptive Text}: Generated using LLMs like GPT-4 to provide detailed overviews and explanations. Additional data can be harvested using web crawlers to extract relevant documentation and forum discussions.
    \item \textit{Executable Code Samples}: Collected from official PyChrono documentation and reliable open-source repositories. 
\end{itemize}

\medskip

\sbelBoldNoIndent{Dataset Components.} The dataset is structured into four JSON files:
\begin{itemize}
    \item \textit{pychrono\_sft\_sim.json}: Instructions and corresponding code outputs for simulation tasks, which is collected using Fig~\ref{fig:sani1}
    \item \textit{pychrono\_sft\_COT.json}: Chain-of-Thought examples with step-by-step reasoning.
    \item \textit{pychrono\_sft\_NL2API.json}: Maps natural language instructions to API calls.
    \item \textit{pychrono\_sft\_API2NL.json}: Maps API calls to descriptive natural language explanations.
\end{itemize}

\medskip

\sbelBoldNoIndent{Structure and Format.} The dataset is structured in JSON format for compatibility with machine learning pipelines. Each entry contains three fields: \textit{instruction}, \textit{input}, and \textit{output}. These fields encapsulate the task description, optional input context, and the expected response. Below are examples from four dataset categories:

\medskip

\sbelItalicNoIndent{1. Simulation Generation}: \texttt{pychrono\_sft\_robot\_sim.json}. This dataset focuses on generating Python code for PyChrono simulations, covering physics settings, model initialization, and visualization.
\vspace{-6pt}
\begin{quote}
\begin{verbatim}
[
  {
    "instruction": "Simulate the Curiosity rover traversing rigid terrain using PyChrono.",
    "input": "Rover model: Curiosity, Terrain: Rigid",
    "output": "import pychrono as chrono\nimport pychrono.robot as robot\n..."
  }
]
\end{verbatim}
\end{quote}

\medskip

\sbelItalicNoIndent{2. Chain-of-Thought}: \texttt{pychrono\_sft\_COT.json}. This dataset includes step-by-step reasoning to enhance the model's ability to logically approach tasks.
\vspace{-6pt}
\begin{quote}
\begin{verbatim}
[
  {
    "instruction": "Explain the steps to build a basic Curiosity rover model in PyChrono.",
    "input": "",
    "output": "1. Import necessary libraries...\n2. Create a simulation system...\n..."
  }
]
\end{verbatim}
\end{quote}

\medskip

\sbelItalicNoIndent{3. Natural Language to API (NL2API) Mapping}: \texttt{pychrono\_sft\_NL2API.json}. This dataset translates natural language instructions into PyChrono API calls.
\begin{quote}
\begin{verbatim}
[
  {
    "instruction": "Generate PyChrono code for a body with material properties and texture.",
    "input": "",
    "output": "ground_mat = chrono.ChContactMaterialNSC()\n..."
  }
]
\end{verbatim}
\end{quote}

\medskip

\sbelItalicNoIndent{4. API to Natural Language Mapping (API2NL) Mapping}: \texttt{pychrono\_sft\_API2NL.json}. This dataset maps API calls to descriptive explanations, aiding comprehension.
\begin{quote}
\begin{verbatim}
[
  {
    "instruction": "Explain the following API: ground_mat = chrono.ChContactMaterialNSC()\n...",
    "input": "",
    "output": "Create a ground body with specified dimensions and material properties."
  }
]
\end{verbatim}
\end{quote}

\sbelBoldNoIndent{Key Features.} The key attributes of this approach for assembling a fine-tuning dataset are as follows:
\begin{itemize}
    \item \textit{Compatibility}: JSON format ensures integration with standard pipelines.
    \item \textit{Diversity}: Covers various PyChrono tasks, e.g., model generation, reasoning, and API2NL/NL2API conversations.
    \item \textit{Scalability}: Can be expanded with new tasks or categories, or by adding more content in the four categories described above.
\end{itemize}

\begin{figure}[!t]
  \centering
  \includegraphics[width=0.9\linewidth]{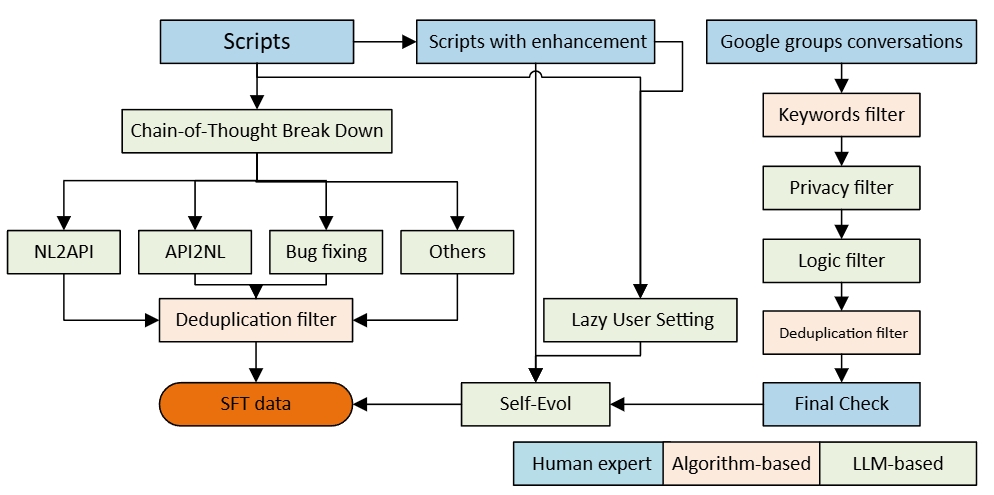}
  \caption{A comprehensive pipeline for synthesizing the Supervised Fine-Tuning (SFT) dataset.}
  \label{fig:syn_pipeline}
\end{figure}

The process of generating the fine-tuning dataset is summarized in Fig. \ref{fig:syn_pipeline}. It begins with the ingestion of PyChrono simulation \textit{Scripts} and \textit{Google group conversations}. The PyChrono Simulation scripts are  moved into a markdown file and subsequently enhanced by expert users who provide sharp comments and clean up the file's content. This is typically time-consuming and expensive. The resulting data undergoes various transformations such as Chain-of-Thought (CoT) breakdown, natural language to API (NL2API) conversion, API to natural language (API2NL) translation, bug fixing, and other enhancements. Google Group conversations are filtered through keyword extraction, privacy checks, and logic validation. Both data streams undergo deduplication and are checked for quality using Lazy User Setting and Self-Evolution mechanisms. Finally, the resulting dataset undergoes a final check before forming the SFT data, leveraging human experts, algorithm-based processes, and LLM-based techniques. The process diagram uses color coding to indicate the source of each component: blue signifies contributions from human experts, green indicates content generated by an LLM, and red denotes outputs from traditional (non-LLM) algorithms.

\section{Experiments}
\label{sec:Numericaltest}
This section compares how several ChronoLLM models, generated based off various third party LLMs, perform in our testing. As a disclaimer, the behavior of a third party LLM in relation to PyChrono tasks is not a direct manifestation of the overall performance of the LLM. If ChronoLLM drawing on LLM A is better than the ChronoLLM drawing on LLM B, this does not imply that A is better than B. It only says that given the data used and our tuning process, the A-based ChronoLLM appears in our testing to be superior to the B-based Chrono LLM. The observation that the A-based ChronoLLM outperforms the B-based ChronoLLM is ultimately subjective, as the final evaluation relies on a separate ``judge'' LLM, which is disucssed in a companion paper \cite{jingquanSimbench2024}.

\subsection{Metrics for Evaluating Digital Twin Generation}
Evaluating the performance of LLMs is non-trivial, particularly in conjunction with code generation tasks \cite{chang2024LLMevaluation, liu2024LLMcodeevaluation}. The evaluation metrics used to assess the quality of an LLM can be categorized into three main types: similarity-based methods, execution-based methods, and LLM-as-judge approaches.

\textbf{Similarity-Based Methods.} These metrics assess the generated output, produced by the LLM in response to a user's prompt, by comparing it to reference code. A commonly used metric is \textbf{BLEU (Bilingual Evaluation Understudy)} \cite{papineni2002bleu1, EVTIKHIEV2023bleu2}, which measures the $n$-gram overlap between the generated and reference texts. \textbf{CodeBLEU} extends BLEU by incorporating syntactic and semantic features specific to source code, thereby providing a more nuanced evaluation of programming tasks. Additionally, the \textbf{ROUGE (Recall-Oriented Understudy for Gisting Evaluation)} family \cite{ganesan2018rouge} focuses on recall-based measures to evaluate the quality of summaries by assessing the overlap of essential components between the generated and reference texts or summaries.

\textbf{Execution-Based Methods.} These metrics evaluate the functional correctness of the generated code by executing it against predefined tests. The metric \textbf{$\text{pass@}k$} \cite{chen2021evaluating1, xu2022evaluating2, liu2024evaluating3} represents the probability that at least one out of $k$ generated samples passes all unit tests, thereby indicating functional accuracy. Another important metric is \textbf{$\text{compile@}k$}, which checks whether the generated software code compiles successfully, thereby ensuring syntactic correctness. These execution-based methods provide a direct measure of whether the generated code performs as intended.

\textbf{LLM-as-a-Judge Approach.} Beyond traditional metrics, the LLM-as-a-judge \cite{zheng2023LLMasJudge} paradigm leverages another LLM to evaluate the quality of the generated code. In a companion paper \cite{jingquanSimbench2024}, we introduced the \textbf{J-LLM} (``J'' from ``Judge''), which utilizes PyChrono reference code and API documentation to assess ChronoLLM's performance. In our experience, J-LLM offers a more reliable evaluation metric compared to similarity-based methods like CodeBLEU and ROUGE-L\textsubscript{Sum}. Furthermore, it exhibits a stronger correlation with $\text{pass@}k$ than with $\text{compile@}k$, suggesting its effectiveness in evaluating both the functional and semantic quality of the generated code. These results, reported in \cite{jingquanSimbench2024}, are obtained in the context of SimBench, which provides a dataset and processes to evaluate the LLM's performance in generating PyChrono digital twins.

In the numerical evaluation section below, all the aforementioned metrics are employed to assess the performance of the LLM in generating simulation scripts. This captures both the syntactic and functional aspects of the generated code.

\subsection{Solutions Compared}
We pursue two distinct approaches to produce ChronoLLMs. The first approach utilizes in-context learning, which leverages well-written API documentation and PyChrono-specific examples to enhance the model's understanding and generation capabilities without altering its underlying parameters -- see subsection \ref{subsec:in_context_learning}. The second approach involves fine-tuning a ChronoLLM model with PyChrono-specific data, allowing the model to adapt its weights. The selected LLMs that anchor ChronoLLMs vary in scale and include both open-source and commercial options allowing for a comprehensive evaluation. Specifically, we apply in-context learning to models from the GPT and LLaMA families. 
In addition, we fine-tune GPT-4o-mini, LLaMA3.1-8B, and LLaMA3.3-70B. This effort facilitates a robust comparison across different LLM architectures and customization methods. 

\subsection{Results for In-Context Learning Efforts}
\label{subsec:res_in_context_learning}

In-context learning (ICL) is one of several tools in the ``prompt engineering'' toolkit. It is a prompt-based method where an LLM leverages provided examples, and is guided to utilize relevant API documentation and contextual instructions at inference time, without any adjustment to its pretrained parameters.  In this exercise, we provided ChronoLLM with structured, detailed examples of PyChrono API use, e.g., library imports, contact and collision settings, visualization configurations, body initialization, joints definitions, and simulation loops. The detailed documentation includes explicit Python code snippets and descriptions of each method's usage, with the goal of enhancing the model's ability to accurately utilize the PyChrono library. The documentation provided for ICL contains:

\begin{itemize}
\item \textbf{Library Imports}: Essential modules required for simulation setup, e.g., \texttt{numpy}, \texttt{scipy}.
\item \textbf{Contact and Collision Settings}: Detailed explanations and examples of initializing systems with Non-Smooth Contact (NSC) or Smooth Contact (SMC) mechanics, configuring contact materials, and defining collision models.
\item \textbf{Visualization Settings}: Instructions for setting up Irrlicht visualization, including window properties, camera positioning, lighting, and visual shape definitions. Information about how to query, at run time, the simulator about state information and numerical solution metrics (amount of time per time step, number of iterations per time step, etc.)
\item \textbf{Body Initialization and Properties}: Examples of creating rigid bodies with specified geometries, mass, inertia, collision settings, and fixed or dynamic states.
\item \textbf{Joints (also called links, in Chrono terminology)}: Examples demonstrating the setup and initialization of various joints (revolute, prismatic, spherical, universal, motor joints) crucial for multibody dynamics. The examples include the definition of the joint type, the joint frame, the joint axes, and the joint limits.
\item \textbf{Simulation Loop}: Structured guidelines for executing the simulation steps and integrating visualization routines.
\end{itemize}

\paragraph{Evaluation metrics.}
Table~\ref{tab:llm_pe_vs_nonpe}, down its last three columns, reports three complementary metrics that together reveal how much each large-language model (LLM) benefits from different forms of context.  
\emph{Score Document} quantifies accuracy when the model sees only the natural-language documentation associated with each task, providing a measure of the model's capacity to reason from prose alone.  
\emph{Score Reference} measures accuracy when the model is given the authoritative reference code but no accompanying documentation, thereby isolating its ability to read and reason about code in the absence of explanatory text.  
\emph{Score Reference Document} evaluates performance when both sources of information are present and thus represents the upper bound under full context.  

In Table \ref{tab:llm_pe_vs_nonpe}, rows highlighted in light red correspond to ICL variants; their performance can be compared directly with the base versions to assess ICL's effectiveness.

\begin{table}[!h]
\centering
\caption{Comparison of LLMs with and without in-context learning.  Rows shaded in {\color{pe}light red} are in-context learning (ICL) variants.}
\label{tab:llm_pe_vs_nonpe}
\begin{tabular}{lccc}
\toprule
\textbf{Test Model} & \ \textbf{J-LLM Doc} & \textbf{J-LLM Ref}& \textbf{J-LLM Ref+Doc} \\
\midrule
llama-3.1-405b-instruct & 38.03 & 33.09 & 39.04 \\
\rowcolor{pe} pe\_llama-3.1-405b-instruct & 41.83 & 34.97 & 40.87 \\
\midrule
deepseek-r1-32b & 34.00 & 33.30 & 39.09 \\
\rowcolor{pe} pe\_deepseek-r1-32b & 46.31 & 35.19 & 40.24 \\
\midrule
llama-3.3-70b-instruct & 41.86 & 33.73 & 41.08 \\
\rowcolor{pe} pe\_llama-3.3-70b-instruct & 48.55 & 35.38 & 40.13 \\
\midrule
llama-3.1-70b-instruct & 39.71 & 33.42 & 39.59 \\
\rowcolor{pe} pe\_llama-3.1-70b-instruct & 43.69 & 34.87 & 39.90 \\
\midrule
llama-3.1-8b-instruct & 34.85 & 31.43 & 37.75 \\
\rowcolor{pe} pe\_llama-3.1-8b-instruct & 37.23 & 33.19 & 39.06 \\
\midrule
llama4\_scout & 37.85 & 33.77 & 40.82 \\
\rowcolor{pe} pe\_llama4\_scout & 48.25 & 34.07 & 39.01 \\
\midrule
llama4\_maverick & 43.57 & 33.07 & 41.20 \\
\rowcolor{pe} pe\_llama4\_maverick & 43.54 & 31.74 & 38.64 \\
\midrule
deepseek-r1-8b & 26.68 & 26.27 & 34.03 \\
\rowcolor{pe} pe\_deepseek-r1-8b & 34.66 & 30.62 & 36.91 \\
\bottomrule
\end{tabular}
\end{table}

\subsection{Results for Fine-Tuning Efforts}
\subsubsection{Pre-trained vs. ICL vs. SFT comparison}
\label{subsubec:PT_ICL_SFT}
GPT-4o, GPT-4o-mini, and Llama3.1-70B were employed in a three-way comparison: pre-trained (shown as ``Pre'' in Table \ref{tab:llm_variants_rows_judge3_filled}) vs. in-context learning (shown as ``ICL'') vs. fine-tuning (shown as ``SFT''). For each of these models, the ICL variant showed  modest improvements in generating contextually correct PyChrono scripts. While the gains were modest, so was the effort of creating the prompts. In general, in-context learning is recommended only in cases where the training data and/or computational resources are limited. Evaluation results for ICL demonstrate noticeable performance variations among different LLM architectures. 

Finally, the results for the SFT models show a significant improvement in performance over both the pre-trained and ICL models.

\begin{table}[!h]
  \centering
  \caption{LLMs across variants (Pretrain / ICL / SFT) and metric families.
  LLM-as-Judge numbers are filled from the provided table; other metrics left blank (—).}
  \label{tab:llm_variants_rows_judge3_filled}
  \begingroup
  \setlength{\tabcolsep}{3.5pt}
  \renewcommand{\arraystretch}{0.95}
  \scriptsize
  \resizebox{\linewidth}{!}{%
  \begin{tabular}{
    l l
    | S[table-format=2.2] S[table-format=2.2]
    | S[table-format=2.2] S[table-format=2.2] S[table-format=2.2]
    | S[table-format=2.2] S[table-format=2.2]
  }
    \toprule
    \multicolumn{2}{c|}{\textbf{Model}} &
    \multicolumn{2}{c|}{\textbf{Similarity-based}} &
    \multicolumn{3}{c|}{\textbf{LLM-as-Judge}} \\
    \cmidrule(lr){1-2}\cmidrule(lr){3-4}\cmidrule(lr){5-7}
    \textbf{LLM} & \textbf{Variant} &
    \textbf{ROUGE-LSUM} & \textbf{CodeBLEU} &
    \textbf{J-LLM Ref+Doc} & \textbf{J-LLM Ref} & \textbf{J-LLM Doc}\\
    \midrule

    \LLM{GPT-4o-mini}
      & Pre & 0.72  &0.61  & 41.80 & 34.46 & 43.22   \\
      & ICL       & 0.76 & 0.61 & 41.22  & 34.92 & 47.09  \\
      & SFT      & 0.96 & 0.92 & 68.03 & 66.94   & 38.03  \\
    \addlinespace

    \LLM{llama-3.3-70b}
      & Pre & 0.73 & 0.60  & 41.08  & 33.73 & 41.86  \\
      & ICL       & 0.74  &0.60  &40.13  & 35.38 & 48.55  \\
      & SFT      &0.82  &0.69  & 43.10 & 37.29 & 38.68 \\
    \addlinespace

    \LLM{llama-3.1-8b}
      & Pre & 0.66 & 0.57 & 37.75 & 31.43  & 34.85 \\
      & ICL       & 0.72 & 0.60 & 39.06 & 33.19 & 37.23   \\
      & SFT      & 0.74 & 0.63 & 38.78 &33.09  & 34.56 \\
    \bottomrule
  \end{tabular}}
  \endgroup
\end{table}

\subsubsection{LoRA results}
\label{subsubsec:LoRA_results}
While full-size SFT can lead to sizable improvements, it is often resource-intensive. This quickly becomes prohibitive for large models that go beyond 70B parameters. In contrast, LoRA offers a more efficient alternative by fine-tuning only a small subset of parameters, significantly reducing the required VRAM, e.g. full-size SFT for the 70B Llama model usually needs more than a single node of eight H100 80GB GPUs, while the LoRA version can reduce the GPU requirements to around four H100 80GB GPUs. This makes LoRA appealing for scenarios with limited GPU resources. In this section, we report the LoRA performance for two Llama3 models at 8B and 70B, and note that LoRA achieves comparable results to the full-size SFT. The LoRA fine-tuning was performed using the same data as the full-size SFT, but different hyperparameters were used to train the LoRA models. Table \ref{tab:lora_results} summarizes the results when comparing LoRA against the pre-trained and SFT models.

\begin{table}[!h]
  \centering
  \caption{LLMs across variants (Pretrain / LoRA / SFT) and metric families.
  LLM-as-Judge numbers are filled from the provided table; other metrics left blank (—).}
  \label{tab:lora_results}
  \begingroup
  \setlength{\tabcolsep}{3.5pt}
  \renewcommand{\arraystretch}{0.95}
  \scriptsize
  \resizebox{\linewidth}{!}{%
  \begin{tabular}{
    l l
    | S[table-format=2.2] S[table-format=2.2]
    | S[table-format=2.2] S[table-format=2.2] S[table-format=2.2]
    | S[table-format=2.2] S[table-format=2.2]
  }
    \toprule
    \multicolumn{2}{c|}{\textbf{Model}} &
    \multicolumn{2}{c|}{\textbf{Similarity-based}} &
    \multicolumn{3}{c|}{\textbf{LLM-as-Judge}} \\
    \cmidrule(lr){1-2}\cmidrule(lr){3-4}\cmidrule(lr){5-7}
    \textbf{LLM} & \textbf{Variant} &
    \textbf{ROUGE-LSUM} & \textbf{CodeBLEU} &
    \textbf{J-LLM Ref+Doc} & \textbf{J-LLM Ref} & \textbf{J-LLM Doc}\\
    \midrule

    \LLM{llama-3.3-70b}
      & Pre & 0.73 & 0.60  & 41.08  & 33.73 & 41.86  \\
      & LoRA       & 0.77  &0.64  &40.38  & 31.64 & 31.62  \\
      & SFT      &0.82  &0.69  & 43.10 & 37.29 & 38.68 \\
    \addlinespace

    \LLM{llama-3.1-8b}
      & Pre & 0.66 & 0.57 & 37.75 & 31.43  & 34.85 \\
      & LoRA       & 0.76 & 0.64 & 39.07 & 32.77 & 34.92   \\
      & SFT      & 0.74 & 0.63 & 38.78 &33.09  & 34.56 \\
    \bottomrule
  \end{tabular}}
  \endgroup
\end{table}
The caveat with LoRA is that, compared to SFT, the latter is generally more robust and easier to train, often requiring fewer epochs and being more tolerant to hyperparameter choices. In contrast, LoRA requires more trial and error to achieve comparable performance.

\subsubsection{Cost of SFT for a small model}
\label{subsubsec:cost_of_fine_tuning}
\begin{figure}[H]
    \centering
    \includegraphics[width=0.75\linewidth]{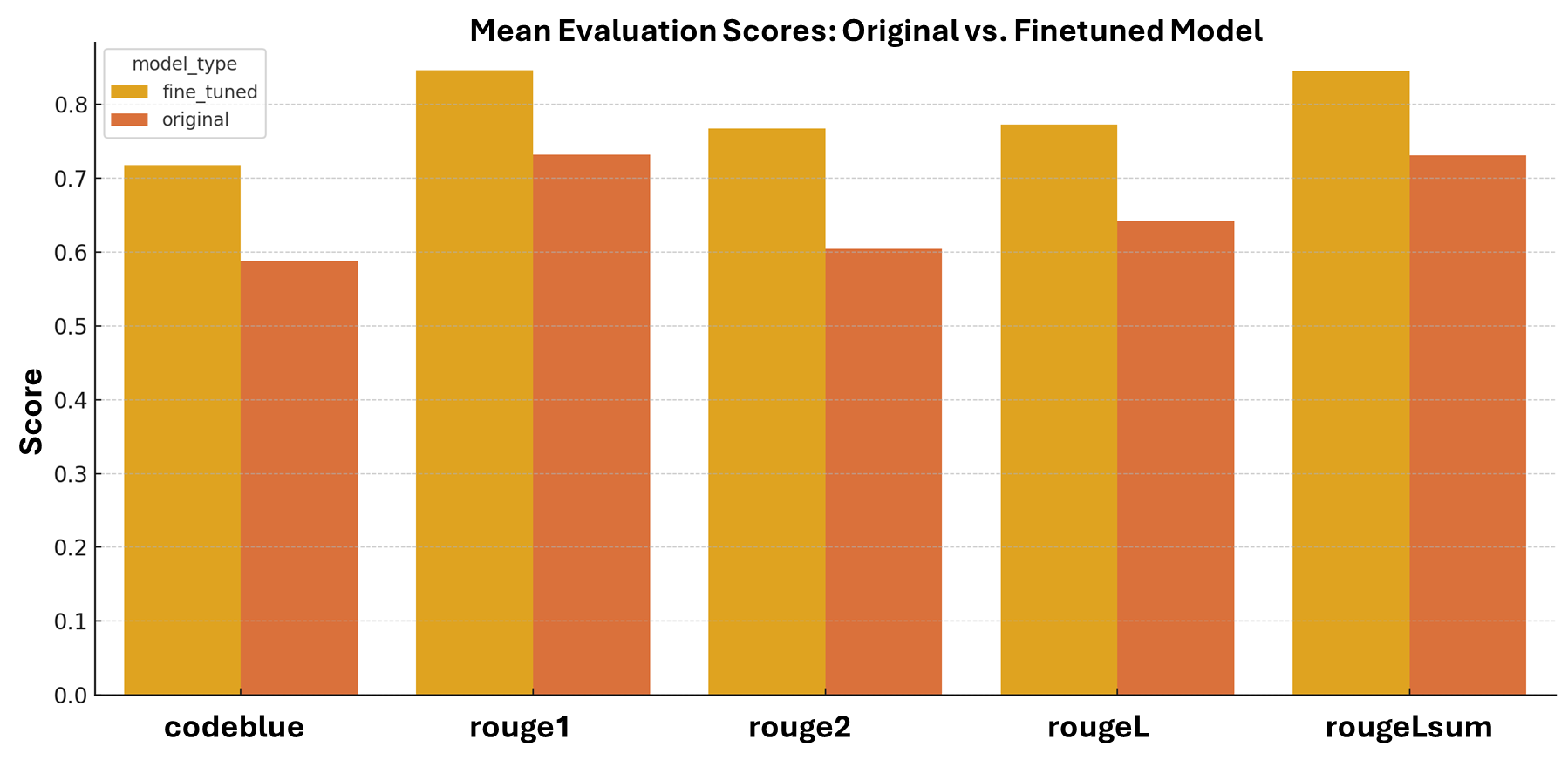}
    \caption{Comparison of CodeBLEU and ROUGE scores for the fine-tuned and pretrained versions of GPT-4.1-mini.}
    \label{fig:fine-tuning-andy}
\end{figure}

For fine-tuning, we close with an example that pertains the fine-tuning of GPT-4.1-mini via OpenAI's API. Once the time-consuming step of data preparation was complete, the entire process took around 20 minutes and came at a cost of roughly \$3 USD. Upon completion of the fine-tuning process, we evaluated the performance of the fine-tuned model using \textbf{CodeBLEU}, which captures code-specific similarity through n-gram overlap, syntax structure, and dataflow consistency, and \textbf{ROUGE}, which measures textual overlap at the unigram, bigram, and sequence levels. These metrics provide both structural and semantic assessments of the generated code's alignment with reference implementations. The fine-tuned version outperformed the pretrained version across all five recorded metrics, see Figure \ref{fig:fine-tuning-andy}. In the figure, \textit{rouge1} and \textit{rouge2} measure the overlap of unigrams and bigrams, respectively, capturing basic content and phrase-level fluency. \textit{rougeL} evaluates the longest common subsequence between generated and reference summaries, reflecting sentence-level structure. We also report \textit{rougeLsum} as an aggregate score that combines the individual ROUGE metrics to provide an overall measure of summarization quality. The results suggest that fine-tuning a closed-source LLM can lead to quantitative improvements in ChronoLLM performance. The process was both quick and inexpensive owing to the small size of the LLM and the fact that the fine-tuning was conducted in the cloud, via the OpenAI's API.

\section{Two Case Studies}
In this section, we present two examples to demonstrate the capabilities of ChronoLLM---one is a simple double arm system (or double pendulum), the other one is a truck. Each example is accompanied by an input prompt, the corresponding generated PyChrono scripts, and frames that we obtained upon executing the PyChrono scripts, i.e., running the simulations. Note that these types of models might be considered ``in-distribution'' for ChronoLLM, as they are generated using data that was qualitatively similar to the one used to train ChronoLLM. We argue that this is not a significant issue. Ultimately, ChronoLLM is meant to assist the users in generating PyChrono scripts -- the users are not interested if their request is in-distribution or not. Ideally, ChronoLLM would be trained on an amount of data that would make any request in-distribution. This is not possible, and we argue that it is not necessary. Even providing a good starting point for the users is a gain.

\subsection{First Case Study: Double Pendulum Simulation}
We start with an easy request to generate a double pendulum -- asking the pretrained LLM to generate the pyChrono script without giving any specific setting of the two bodies, e.g. initial position, length, or mass. This will test whether the \textit{pretrained} model can generate a most basic model. The entire ChronoLLM response is shown in Figure \ref{fig:gpt4o-mini-double-pendulum}. 

\paragraph{Prompt:}
\begin{quote}
\textit{You are a PyChrono expert. Please give me a simulation script for the double pendulum in PyChrono.}
\end{quote}
\begin{figure*}[ht]
	\centering
	\ovalbox{
		\begin{minipage}{1.0\linewidth}
			{\tiny
				\begin{lstlisting}[breaklines=true, numbers=none]
Creating a double pendulum simulation using PyChrono involves setting up the environment, defining the pendulum components, and specifying their interactions. Below is a script that illustrates how to simulate a double pendulum using PyChrono.

Make sure you have PyChrono installed. If you haven't installed it yet, you can do so via pip:

```bash
pip install pychrono
```

Here's a simple script to simulate a double pendulum:

```python
import pychrono as chrono

# Create the Chrono system

system = chrono.ChSystemNSC()

# Define parameters for the pendulum
length1 = 1.0  # Length of the first arm
length2 = 1.0  # Length of the second arm
mass1 = 1.0    # Mass of the first bob
mass2 = 1.0    # Mass of the second bob

# Create the first pendulum body (bob)
bob1 = chrono.ChBodyAuxRef()
bob1.SetMass(mass1)
bob1.SetInertiaXX(chrono.ChVectorD(0.1, 0.1, 0.1))  # Set some inertia
bob1.SetPos(chrono.ChVectorD(length1, length1, 0))
system.Add(bob1)

# Create the second pendulum body (bob)
bob2 = chrono.ChBodyAuxRef()
bob2.SetMass(mass2)
bob2.SetInertiaXX(chrono.ChVectorD(0.1, 0.1, 0.1))  # Set some inertia
bob2.SetPos(chrono.ChVectorD(length1 + length2, length1 + length2, 0))
system.Add(bob2)

# Create the first joint (revolute) for the first pendulum
joint1 = chrono.ChLinkLockRevolute()
joint1.Initialize(bob1, system.GetGround(), chrono.ChVectorD(0, 0, 0), chrono.ChVectorD(0, 1, 0))
system.Add(joint1)

# Create the second joint (revolute) for the second pendulum
joint2 = chrono.ChLinkLockRevolute()
joint2.Initialize(bob1, bob2, chrono.ChVectorD(length1, length1, 0), chrono.ChVectorD(0, 1, 0))
system.Add(joint2)

# Initial conditions
bob1.SetWvel_par(chrono.ChVectorD(0, 0, 0))  # Initial angular velocity for bob1
bob2.SetWvel_par(chrono.ChVectorD(0, 0, 0))  # Initial angular velocity for bob2

# Set up the simulation parameters
step_size = 0.01
end_time = 10.0

# Fill history arrays to visualize later
positions1 = []
positions2 = []

# Simulation loop
while system.GetChTime() < end_time:
    system.DoStepDynamics(step_size)

    # Record the positions of the bobs
    positions1.append(bob1.GetPos())
    positions2.append(bob2.GetPos())

# Visualization (optional)
import matplotlib.pyplot as plt
import numpy as np

# Extract x and y positions
x1 = [pos.x for pos in positions1]
y1 = [pos.y for pos in positions1]
x2 = [pos.x for pos in positions2]
y2 = [pos.y for pos in positions2]

# Plot the trajectory of the bobs
plt.figure()
plt.plot(x1, y1, label='Bob 1 Trajectory', color='blue')
plt.plot(x2, y2, label='Bob 2 Trajectory', color='red')
plt.title('Double Pendulum Motion')
plt.xlabel('X Position (m)')
plt.ylabel('Y Position (m)')
plt.legend()
plt.axis('equal')
plt.grid()
plt.show()
```

### Explanation:

1. **System Initialization**: The script initializes the PyChrono system.
2. **Pendulum Parameters**: You define the lengths and masses of the pendulums.
3. **Bodies Creation**: Two bobs (bodies) are created with mass and inertia.
4. **Joints**: Two revolute joints allow the bobs to swing.
5. **Initial Velocities**: You can set the initial angular velocities if needed.
6. **Simulation Loop**: The main loop where the system is advanced through time, collecting positions for visualization.
7. **Visualization**: After the simulation, the positions are plotted using `matplotlib`.

This script should give you a basic setup for simulating a double pendulum. You can customize the masses, lengths, and initial conditions as needed.
				\end{lstlisting}
			}
		\end{minipage}
	}
	\caption{The generated response of Pretrained GPT4o-mini for the PyChrono double pendulum simulation.}
	\label{fig:gpt4o-mini-double-pendulum}
\end{figure*}

Even though the model's output appears plausible at a first glance, a careful inspection reveals numerous conceptual and API errors that would prevent execution. Even if those were fixed, the PyChrono script would produce non-physical motion. Some of the issues noted are as follows:
\begin{enumerate}
\item \textbf{API family mixing and hallucinations.} The code blends legacy and modern Chrono APIs (e.g., \texttt{ChVectorD} vs.\ \texttt{ChVector3d}) and calls non-existent methods such as \texttt{system.GetGround()} and \texttt{SetWvel\_par()}.
\item \textbf{Incorrect joint initialization.} \texttt{ChLinkLockRevolute.Initialize} requires a full frame (position + orientation); supplying raw vectors is semantically wrong and brittle across versions.
\item \textbf{World modeling errors.} No explicit fixed ground body is created and gravity is left implicit, making anchor references ill-defined and results non-reproducible.
\item \textbf{Frame/axis specification mistakes.} The revolute axis is inconsistent with planar motion (should be $Z$ for XY-plane dynamics), and anchor orientations are omitted, leaving joint axes underdetermined.
\item \textbf{Coordinate/units ambiguity.} Axis conventions are undocumented and unit scaling is unclear, producing questionable force/time scales.
\end{enumerate}

\begin{figure}
    \centering
    \includegraphics[width=0.8\textwidth]{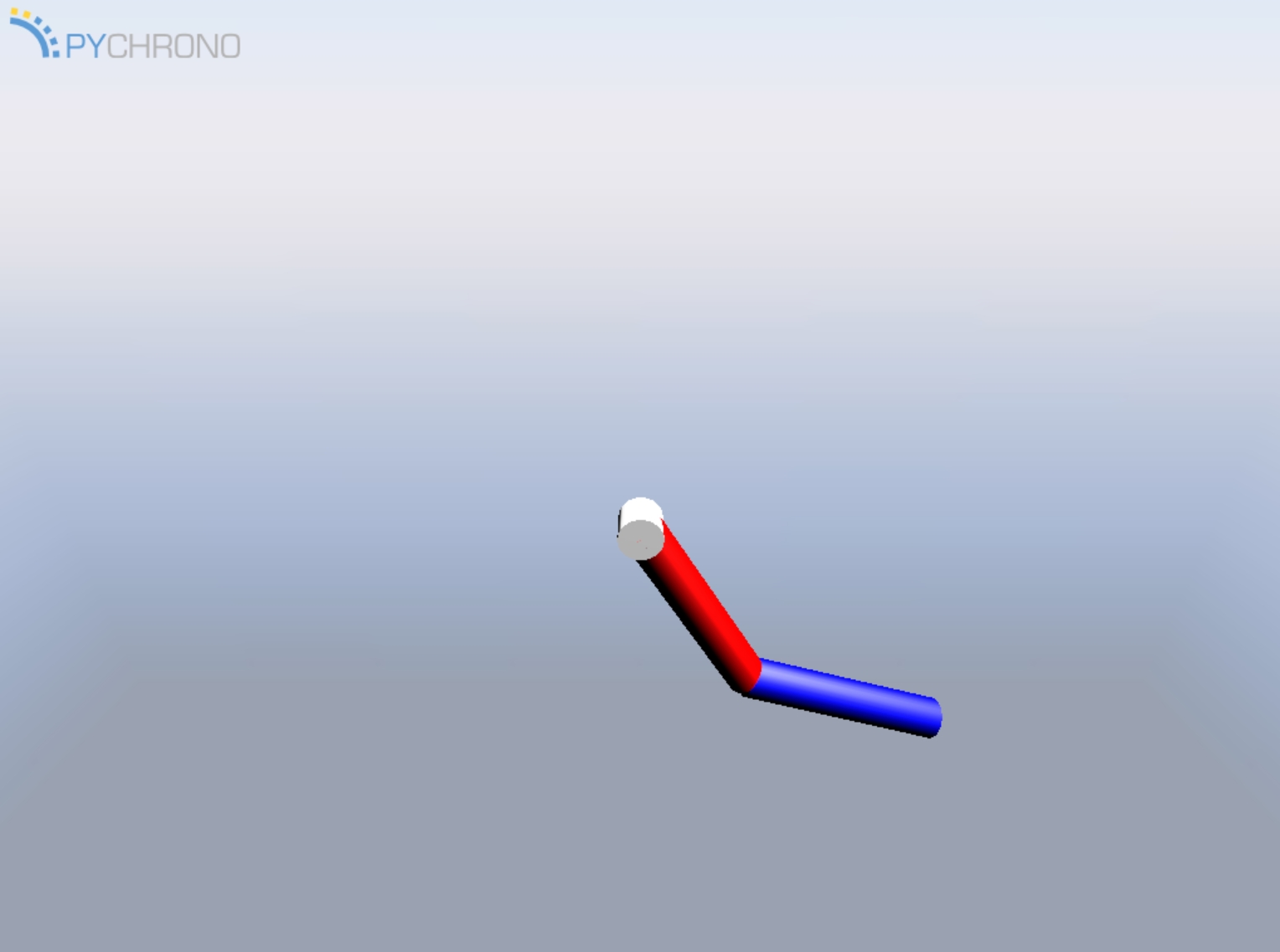}
    \caption{Visualization of the double pendulum simulation in PyChrono.}
    \label{fig:double_pendulum_simulation}
\end{figure}

The same prompt used with ChronoLLM (fine-tuned off ChatGPT4o-mini) generates the PyChrono script shown in Figs. \ref{fig:sft-gpt4o-mini-double-pendulum-1}~and~\ref{fig:sft-gpt4o-mini-double-pendulum-2}, which can be executed without any error. A snapshot of the simulation is shown in Figure \ref{fig:double_pendulum_simulation}.

\begin{figure*}[ht]
	\centering
	\ovalbox{
		\begin{minipage}{1.0\linewidth}
			{\tiny
				\begin{lstlisting}[breaklines=true, numbers=none]
Certainly! Here's a simple script to simulate a double pendulum using PyChrono. Save this code in a file with a `.py` extension and run it using Python.

```python
import pychrono.core as chrono
import pychrono.irrlicht as chronoirr
import math

# Initialize the Chrono simulation system
sys = chrono.ChSystemNSC()

# Set the gravitational acceleration for the system (in m/s^2)
sys.SetGravitationalAcceleration(chrono.ChVector3d(0, -9.81, 0))  # g = 9.81 m/s^2

# Create the ground body and add it to the simulation
ground = chrono.ChBody()
sys.Add(ground)
ground.SetFixed(True)  # The ground is fixed and does not move
ground.EnableCollision(False)  # Disable collision detection for the ground

# Add a visualization cylinder to the ground
cyl_1 = chrono.ChVisualShapeCylinder(0.2, 0.4)  # Cylinder with radius 0.2 and height 0.4
ground.AddVisualShape(cyl_1, chrono.ChFramed(chrono.ChVector3d(0, 0, +1)))

# Create the first pendulum body and add it to the simulation
pend_1 = chrono.ChBody()
sys.AddBody(pend_1)
pend_1.SetFixed(False)  # The pendulum can move
pend_1.EnableCollision(False)  # Disable collision detection for the pendulum
pend_1.SetMass(1)  # Set the mass of the pendulum (in kg)
pend_1.SetInertiaXX(chrono.ChVector3d(0.2, 1, 1))  # Set the inertia tensor (in kg m^2)

# Add a visualization cylinder to the first pendulum
cyl_1 = chrono.ChVisualShapeCylinder(0.2, 2)  # Cylinder with radius 0.2 and height 2
pend_1.AddVisualShape(cyl_1, chrono.ChFramed(chrono.VNULL, chrono.QuatFromAngleY(chrono.CH_PI_2)))

# Set the initial position of the first pendulum (center of mass) in the absolute frame
# The pendulum is initially horizontal, pointing towards the positive X direction
pend_1.SetPos(chrono.ChVector3d(1, 0, 1))

# Create a revolute hinge to allow the first pendulum to swing
hinge_1 = chrono.ChLinkLockRevolute()
hinge_1.Initialize(ground, pend_1, chrono.ChFramed(chrono.ChVector3d(0, 0, 1), chrono.ChQuaterniond(1, 0, 0, 0)))
sys.AddLink(hinge_1)

# Create the second pendulum body and add it to the simulation
pend_2 = chrono.ChBody()
sys.AddBody(pend_2)
pend_2.SetFixed(False)  # The pendulum can move
pend_2.EnableCollision(False)  # Disable collision detection for the pendulum
pend_2.SetMass(1)  # Set the mass of the pendulum (in kg)
pend_2.SetInertiaXX(chrono.ChVector3d(0.2, 1, 1))  # Set the inertia tensor (in kg m^2)

# Add a visualization cylinder to the second pendulum
cyl_2 = chrono.ChVisualShapeCylinder(0.2, 2)  # Cylinder with radius 0.2 and height 2
pend_2.AddVisualShape(cyl_2, chrono.ChFramed(chrono.VNULL, chrono.QuatFromAngleY(chrono.CH_PI_2)))

# Set the initial position of the second pendulum (center of mass) in the absolute frame
# The second pendulum is initially horizontal, pointing towards the positive X direction
pend_2.SetPos(chrono.ChVector3d(3, 0, 1))

# Create a revolute hinge to allow the second pendulum to swing
hinge_2 = chrono.ChLinkLockRevolute()
hinge_2.Initialize(pend_1, pend_2, chrono.ChFramed(chrono.ChVector3d(2, 0, 1), chrono.ChQuaterniond(1, 0, 0, 0)))
sys.AddLink(hinge_2)

# Create the Irrlicht visualization system
vis = chronoirr.ChVisualSystemIrrlicht()
vis.AttachSystem(sys)  # Attach the Chrono system to the visualization
vis.SetWindowSize(1024, 768)  # Set the window size for the visualization
vis.SetWindowTitle('Double Pendulum Simulation')  # Set the window title
vis.Initialize()  # Initialize the visualization system
vis.AddLogo(chrono.GetChronoDataFile('logo_pychrono_alpha.png'))  # Add a logo to the visualization
vis.AddSkyBox()  # Add a skybox for better visual appearance
vis.AddCamera(chrono.ChVector3d(0, 3, 6))  # Add a camera to the visualization
vis.AddTypicalLights()  # Add typical lights for better visualization

# Simulation loop
log_info = True  # Flag to control logging
while vis.Run():
    vis.BeginScene()  # Begin the scene for rendering
    vis.Render()  # Render the scene
    vis.EndScene()  # End the scene
    sys.DoStepDynamics(1e-2)  # Advance the simulation by a time step of 0.001 seconds

    # Log information after 1 second of simulation time
    if log_info and sys.GetChTime() > 1:
        pos_1 = pend_1.GetPos()  # Get the position of the first pendulum
        pos_2 = pend_2.GetPos()  # Get the position of the second pendulum
        print("t = ", sys.GetChTime())
        print("Pendulum 1: ", pos_1.x, "  ", pos_1.y)
        print("Pendulum 2: ", pos_2.x, "  ", pos_2.y)
        lin_vel_1 = pend_1.GetPosDt()  # Get the linear velocity of the first pendulum
        lin_vel_2 = pend_2.GetPosDt()  # Get the linear velocity of the second pendulum
        print("Pendulum 1 velocity: ", lin_vel_1.x, "  ", lin_vel_1.y)
        print("Pendulum 2 velocity: ", lin_vel_2.x, "  ", lin_vel_2.y)
        log_info = False  # Disable further logging

```
				\end{lstlisting}
			}
		\end{minipage}
	}
	\caption{The generated response of finetuned GPT4o-mini for the PyChrono double pendulum simulation.(Part 1 of 2).}
	\label{fig:sft-gpt4o-mini-double-pendulum-1}
\end{figure*}

\begin{figure*}[ht]
	\centering
	\ovalbox{
		\begin{minipage}{1.0\linewidth}
			{\tiny
				\begin{lstlisting}[breaklines=true, numbers=none]

### Explanation:
1. **System Setup**: The simulation starts by initializing the Chrono system with the `ChSystemNSC()` class, which handles the physics calculations. Gravity is set to 9.81 m/s^2 to simulate real-world conditions.

2. **Ground Creation**: A fixed ground body is added to the system, which acts as the stationary point for the pendulums. A visualization element (cylinder) is added to represent the ground.

3. **First Pendulum**: A pendulum body is created, which is not fixed, allowing it to swing. It is initialized horizontally with a mass of 1 kg and inertia properties. A visualization element is added to it.

4. **Hinge Setup**: A revolute hinge is created to allow the first pendulum to swing. It connects the first pendulum to the ground.

5. **Second Pendulum**: A second pendulum is created similarly to the first, with a mass of 1 kg and inertia properties. It is also visualized with a cylinder.

6. **Hinge Between Pendulums**: A revolute hinge is created between the first and second pendulum, allowing the second pendulum to swing from the first.

7. **Visualization Setup**: An Irrlicht visualization system is attached to provide real-time rendering of the simulation. A camera and lighting setup enhance the visual representation.

8. **Simulation Loop**: The simulation runs in a loop where the system is updated every 0.01 seconds. The rendering is done in each step to display the current state of the pendulums. Logging is performed after 1 second to output the position and velocity of both pendulums.

This explanation breaks down the key steps in setting up and running a double pendulum simulation in PyChrono.
				\end{lstlisting}
			}
		\end{minipage}
	}
	\caption{The generated response of finetuned GPT4o-mini for the PyChrono double pendulum simulation.(Part 2 of 2).}
	\label{fig:sft-gpt4o-mini-double-pendulum-2}
\end{figure*}

\FloatBarrier

\subsection{Second Case Study: MAN 10T Truck}
The Maschinenfabrik Augsburg-N\"urnberg (MAN) 10T truck is a high-mobility, all-wheel-drive military truck used for transporting material in off-road conditions. The example demonstrates the capabilities of ChronoLLM in generating complex simulations involving systems with multiple bodies, constraints, tires, powertrain, and more. The input prompt for this example is as follows:
\paragraph{Prompt:}
\begin{quote}
\textit{Create a PyChrono simulation showcasing the dynamics of a MAN 10t truck driving on a rigid terrain. The system includes a vehicle model with configurable visualization and collision settings, a TMEASY tire model, and real-time driver controls for steering, throttle, and braking. Visualize the simulation using the Irrlicht visualization system with a chase camera, directional lighting, a skybox, and customizable terrain textures and logos.}
\end{quote}

\begin{figure}
    \centering
    \includegraphics[width=0.8\textwidth]{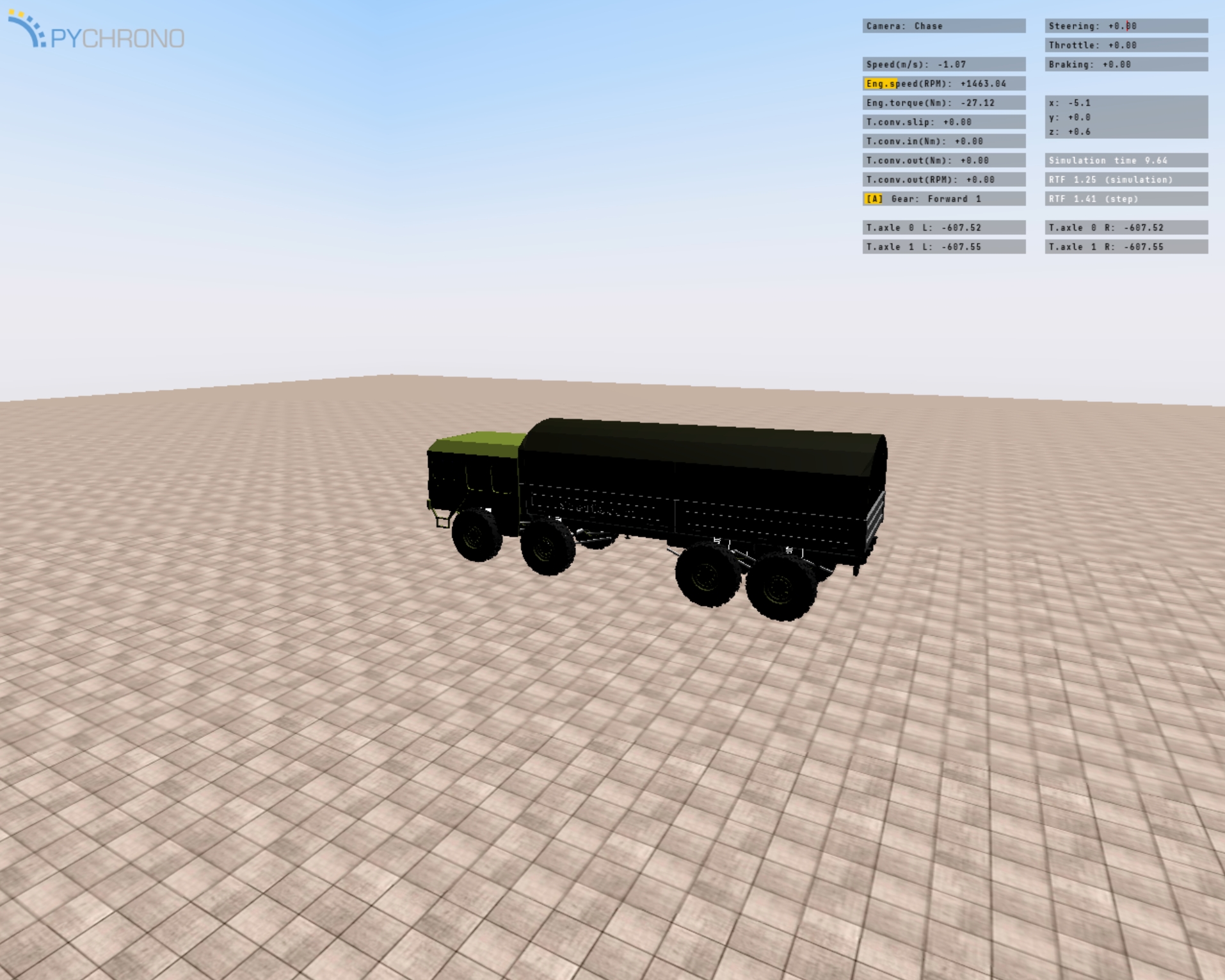}
    \caption{Visualization of the MAN truck simulation in PyChrono.}
    \label{fig:man_10t_truck}
\end{figure}

The PyChrono script generated by ChronoLLM using finetuned GPT4o-mini is shown in Figure \ref{fig:sft-gpt4o-mini-man-10t-truck}. The generated code is a complete simulation setup using the PyChrono library, which includes the creation of a physical system, the addition of a truck model, and the implementation of a visualization system. A snapshot of the simulation is shown in Figure \ref{fig:man_10t_truck}. Subsequent modifications to the script might be needed to get a particular truck model, but this script provides a good starting point to that end.

\begin{figure*}[ht]
	\centering
	\ovalbox{
		\begin{minipage}{1.0\linewidth}
			{\tiny
				\begin{lstlisting}[breaklines=true, numbers=none]
import pychrono.core as chrono
import pychrono.irrlicht as irr
import pychrono.vehicle as veh
import math

chrono.SetChronoDataPath(chrono.GetChronoDataPath())
veh.SetDataPath(chrono.GetChronoDataPath() + 'vehicle/')

# Initial vehicle location and orientation
initLoc = chrono.ChVector3d(0, 0, 0.5)
initRot = chrono.ChQuaterniond(1, 0, 0, 0)
# Visualization type for vehicle parts (PRIMITIVES, MESH, or NONE)
vis_type = veh.VisualizationType_PRIMITIVES
vis_type_mesh = veh.VisualizationType_MESH
# Collision type for chassis (PRIMITIVES, MESH, or NONE)
chassis_collision_type = veh.CollisionType_NONE
# Type of tire model (RIGID, TMEASY)
tire_model = veh.TireModelType_TMEASY
# Rigid terrain
# terrain_model = veh.RigidTerrain.BOX
terrainHeight = 0      # terrain height
terrainLength = 100.0  # size in X direction
terrainWidth = 100.0   # size in Y direction
# Poon chassis tracked by the camera
trackPoint = chrono.ChVector3d(-3.0, 0.0, 1.1)
# Contact method
contact_method = chrono.ChContactMethod_NSC
contact_vis = False
# Simulation step sizes
step_size = 1e-3
tire_step_size = step_size
# Time interval between two render frames
render_step_size = 1.0 / 50  # FPS = 50

# Create the MAN vehicle, set parameters, and initialize
vehicle = veh.MAN_10t()
vehicle.SetContactMethod(contact_method)
vehicle.SetChassisCollisionType(chassis_collision_type)
vehicle.SetChassisFixed(False)
vehicle.SetInitPosition(chrono.ChCoordsysd(initLoc, initRot))
vehicle.SetTireType(tire_model)
vehicle.SetTireStepSize(tire_step_size)
vehicle.Initialize()
vehicle.SetChassisVisualizationType(vis_type_mesh)
vehicle.SetSuspensionVisualizationType(vis_type)
vehicle.SetSteeringVisualizationType(vis_type)
vehicle.SetWheelVisualizationType(vis_type_mesh)
vehicle.SetTireVisualizationType(vis_type_mesh)
vehicle.GetSystem().SetCollisionSystemType(chrono.ChCollisionSystem.Type_BULLET)
# Create the terrain
patch_mat = chrono.ChContactMaterialNSC()
patch_mat.SetFriction(0.9)
patch_mat.SetRestitution(0.01)
terrain = veh.RigidTerrain(vehicle.GetSystem())
patch = terrain.AddPatch(patch_mat,
    chrono.ChCoordsysd(chrono.ChVector3d(0, 0, 0), chrono.QUNIT),
    terrainLength, terrainWidth)
patch.SetTexture(veh.GetDataFile("terrain/textures/tile4.jpg"), 200, 200)
patch.SetColor(chrono.ChColor(0.8, 0.8, 0.5))
terrain.Initialize()
# Create the vehicle Irrlicht interface
vis = veh.ChWheeledVehicleVisualSystemIrrlicht()
vis.SetWindowTitle('MAN 10t Demo')
vis.SetWindowSize(1280, 1024)
vis.SetChaseCamera(trackPoint, 15.0, 0.5)
vis.Initialize()
vis.AddLogo(chrono.GetChronoDataFile('logo_pychrono_alpha.png'))
vis.AddLightDirectional()
vis.AddSkyBox()
vis.AttachVehicle(vehicle.GetVehicle())
# Create the driver system
driver = veh.ChInteractiveDriverIRR(vis)
# Set the time response for steering and throttle keyboard inputs.
steering_time = 1.0  # time to go from 0 to +1 (or from 0 to -1)
throttle_time = 1.0  # time to go from 0 to +1
braking_time = 0.3   # time to go from 0 to +1
driver.SetSteeringDelta(render_step_size / steering_time)
driver.SetThrottleDelta(render_step_size / throttle_time)
driver.SetBrakingDelta(render_step_size / braking_time)
driver.Initialize()

# output vehicle mass
print( "VEHICLE MASS: ",  vehicle.GetVehicle().GetMass())
# Number of simulation steps between miscellaneous events
render_steps = math.ceil(render_step_size / step_size)
# Initialize simulation frame counter s
realtime_timer = chrono.ChRealtimeStepTimer()
step_number = 0
render_frame = 0
while vis.Run() :
    time = vehicle.GetSystem().GetChTime()
    # Render scene and output POV-Ray data
    if (step_number % render_steps == 0) :
        vis.BeginScene()
        vis.Render()
        vis.EndScene()
        render_frame += 1
    # Get driver inputs
    driver_inputs = driver.GetInputs()
    # Update modules (process inputs from other modules)
    driver.Synchronize(time)
    terrain.Synchronize(time)
    vehicle.Synchronize(time, driver_inputs, terrain)
    vis.Synchronize(time, driver_inputs)

    # Advance simulation for one timestep for all modules
    driver.Advance(step_size)
    terrain.Advance(step_size)
    vehicle.Advance(step_size)
    vis.Advance(step_size)
    # Increment frame number
    step_number += 1

    # Spin in place for real time to catch up
    realtime_timer.Spin(step_size)
				\end{lstlisting}
			}   
		\end{minipage}
	}
	\caption{The generated code of finetuned GPT4o-mini for the PyChrono MAN truck simulation.}
	\label{fig:sft-gpt4o-mini-man-10t-truck}
\end{figure*}

\begin{figure}
            \centering
            \includegraphics[width=0.8\textwidth]{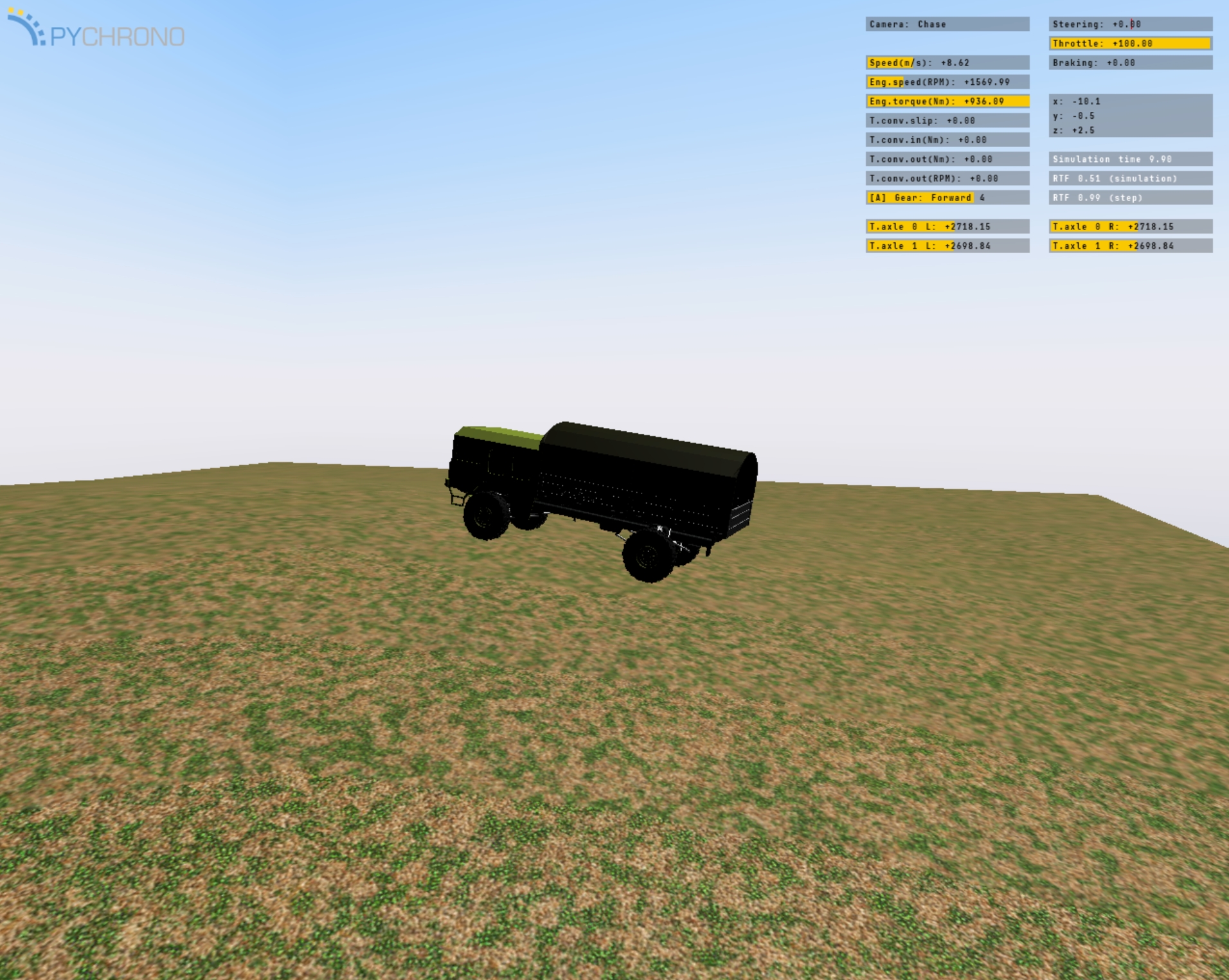}
            \caption{Snapshot of the MAN 5t truck simulation.}
            \label{fig:man_5t_visualization}
\end{figure}

Beyond the first prompt request, it is often the case that the user would like to modify the generated script to get a particular truck model, a different terrain, a different perspective in visualization, etc. This is the second step of this experiment---the model is extended with through another prompt request, which reads as shown below. The script generated is shown in Fig. \ref{fig:sft-gpt4o-mini-man-truck-2}. A snapshot of the simulation is shown in Figure \ref{fig:man_5t_visualization}. Note that MAN 10t has 10 wheels, while MAN 5t has four wheels.
\begin{quote}
\textit{Make the following changes please: Initial vehicle location changed from \texttt{(0, 0, 0.5)} to \texttt{(-20, 0, 1.5)}. Vehicle type changed from \texttt{veh.MAN\_10t} to \texttt{veh.MAN\_5t}. Terrain changed from a flat rigid terrain to a terrain of rigid hills with a height map. Terrain texture changed from \texttt{"tile4.jpg"} to \texttt{"grass.jpg"}.}
\end{quote}

\begin{figure*}[ht]
	\centering
	\ovalbox{
		\begin{minipage}{1.0\linewidth}
			{\tiny
				\begin{lstlisting}[breaklines=true, numbers=none]
import pychrono.core as chrono
import pychrono.irrlicht as irr
import pychrono.vehicle as veh
import math

chrono.SetChronoDataPath(chrono.GetChronoDataPath())
veh.SetDataPath(chrono.GetChronoDataPath() + 'vehicle/')

# Initial vehicle location and orientation
initLoc = chrono.ChVector3d(-20, 0, 1.5)
initRot = chrono.ChQuaterniond(1, 0, 0, 0)
# Visualization type for vehicle parts (PRIMITIVES, MESH, or NONE)
vis_type = veh.VisualizationType_PRIMITIVES
vis_type_mesh = veh.VisualizationType_MESH
# Collision type for chassis (PRIMITIVES, MESH, or NONE)
chassis_collision_type = veh.CollisionType_NONE
# Type of tire model (RIGID, TMEASY)
tire_model = veh.TireModelType_TMEASY
# Rigid terrain
# terrain_model = veh.RigidTerrain.BOX
terrainHeight = 0      # terrain height
terrainLength = 100.0  # size in X direction
terrainWidth = 100.0   # size in Y direction
# Poon chassis tracked by the camera
trackPoint = chrono.ChVector3d(-3.0, 0.0, 1.1)
# Contact method
contact_method = chrono.ChContactMethod_NSC
contact_vis = False
# Simulation step sizes
step_size = 1e-3
tire_step_size = step_size
# Time interval between two render frames
render_step_size = 1.0 / 50  # FPS = 50

# Create the MAN 5t vehicle, set parameters, and initialize
vehicle = veh.MAN_5t() 
vehicle.SetContactMethod(contact_method)
vehicle.SetChassisCollisionType(chassis_collision_type)
vehicle.SetChassisFixed(False)
vehicle.SetInitPosition(chrono.ChCoordsysd(initLoc, initRot))
vehicle.SetTireType(tire_model)
vehicle.SetTireStepSize(tire_step_size)
vehicle.Initialize()
vehicle.SetChassisVisualizationType(vis_type_mesh)
vehicle.SetSuspensionVisualizationType(vis_type)
vehicle.SetSteeringVisualizationType(vis_type)
vehicle.SetWheelVisualizationType(vis_type_mesh)
vehicle.SetTireVisualizationType(vis_type_mesh)
vehicle.GetSystem().SetCollisionSystemType(chrono.ChCollisionSystem.Type_BULLET)

# Create the terrain of rigid hills with a height map
patch_mat = chrono.ChContactMaterialNSC()
patch_mat.SetFriction(0.9)
patch_mat.SetRestitution(0.01)
terrain = veh.RigidTerrain(vehicle.GetSystem())
patch = terrain.AddPatch(patch_mat, chrono.ChCoordsysd(chrono.ChVector3d(0, 0, 0), chrono.QUNIT),veh.GetDataFile("terrain/height_maps/bump64.bmp"), 64.0, 64.0, 0.0, 3.0)
patch.SetTexture(veh.GetDataFile("terrain/textures/grass.jpg"), 6, 6)
patch.SetColor(chrono.ChColor(0.8, 0.8, 0.5))
terrain.Initialize()

# Create the vehicle Irrlicht interface
vis = veh.ChWheeledVehicleVisualSystemIrrlicht()
vis.SetWindowTitle('MAN 5t Demo')
vis.SetWindowSize(1280, 1024)
vis.SetChaseCamera(trackPoint, 15.0, 0.5)
vis.Initialize()
vis.AddLogo(chrono.GetChronoDataFile('logo_pychrono_alpha.png'))
vis.AddLightDirectional()
vis.AddSkyBox()
vis.AttachVehicle(vehicle.GetVehicle())

# Create the driver system
driver = veh.ChInteractiveDriverIRR(vis)
# Set the time response for steering and throttle keyboard inputs.
steering_time = 1.0  # time to go from 0 to +1 (or from 0 to -1)
throttle_time = 1.0  # time to go from 0 to +1
braking_time = 0.3   # time to go from 0 to +1
driver.SetSteeringDelta(render_step_size / steering_time)
driver.SetThrottleDelta(render_step_size / throttle_time)
driver.SetBrakingDelta(render_step_size / braking_time)
driver.Initialize()

# output vehicle mass
print( "VEHICLE MASS: ",  vehicle.GetVehicle().GetMass())
# Number of simulation steps between miscellaneous events
render_steps = math.ceil(render_step_size / step_size)
# Initialize simulation frame counter s
realtime_timer = chrono.ChRealtimeStepTimer()
step_number = 0
render_frame = 0
while vis.Run() :
    time = vehicle.GetSystem().GetChTime()
    # Render scene and output POV-Ray data
    if (step_number % render_steps == 0) :
        vis.BeginScene()
        vis.Render()
        vis.EndScene()
        render_frame += 1
    # Get driver inputs
    driver_inputs = driver.GetInputs()
    # Update modules (process inputs from other modules)
    driver.Synchronize(time)
    terrain.Synchronize(time)
    vehicle.Synchronize(time, driver_inputs, terrain)
    vis.Synchronize(time, driver_inputs)
    # Advance simulation for one timestep for all modules
    driver.Advance(step_size)
    terrain.Advance(step_size)
    vehicle.Advance(step_size)
    vis.Advance(step_size)
    # Increment frame number
    step_number += 1
    # Spin in place for real time to catch up
    realtime_timer.Spin(step_size)
				\end{lstlisting}
			}   
		\end{minipage}
	}
\caption{The generated code of finetuned GPT4o-mini for the second round modifications of the PyChrono MAN 5t truck simulation.}
\label{fig:sft-gpt4o-mini-man-truck-2}
\end{figure*}

\begin{figure}[h!]
    \centering
    \begin{subfigure}{0.48\textwidth}
        \centering
        \includegraphics[width=\linewidth]{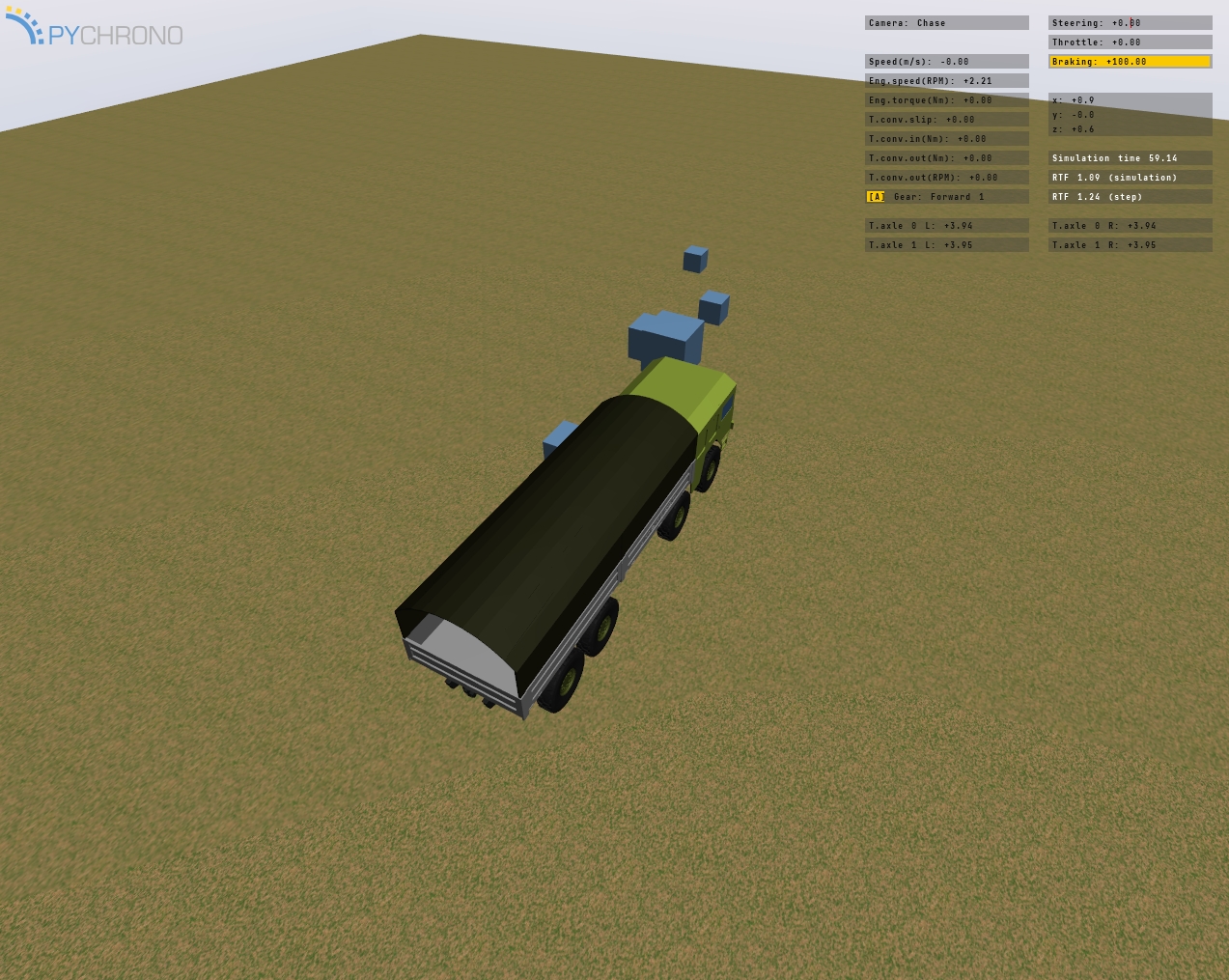}
        \caption{visualization of the Man truck simulation with third round modifications}
        \label{fig:lidar_sim_view1}
    \end{subfigure}
    \hfill
    \begin{subfigure}{0.48\textwidth}
        \centering
        \includegraphics[width=\linewidth]{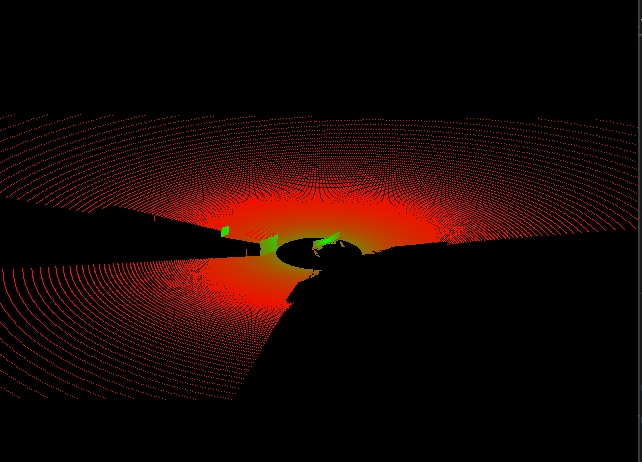}
        \caption{visualization of the generated Lidar view for the MAN truck}
        \label{fig:lidar_sim_view2}
    \end{subfigure}

    \caption{Visualization of the third prompt: a snapshot of the vehicle, and a rendering of what the LiDAR model in the vehicle picks up during the simulation.}
    \label{fig:lidar_sensor_simulation}
\end{figure}

One might want to extend the model with a third round of modifications, e.g., adding sensor to the simulation to simulate the real user experience to interact with LLM. The prompt request is shown below. The script generated is shown in Fig. \ref{fig:sft-gpt4o-mini-man-truck-3-2}. A snapshot of the simulation is shown in Figure \ref{fig:lidar_sensor_simulation}.
\begin{quote}
\textit{Imported pychrono.sensor and numpy modules. Added sensor manager and lidar sensor. Created and added random boxes to the simulation. Terrain texture changed from "tile4.jpg" to "grass.jpg". Updated sensor manager within the simulation loop.}
\end{quote}

\begin{figure*}[ht]
	\centering
	\ovalbox{
		\begin{minipage}{1.0\linewidth}
			{\tiny
				\begin{lstlisting}[breaklines=true, numbers=none]
import pychrono.core as chrono
import pychrono.irrlicht as irr
import pychrono.vehicle as veh
import pychrono.sensor as sens
import numpy as np
import math

chrono.SetChronoDataPath(chrono.GetChronoDataPath())
veh.SetDataPath(chrono.GetChronoDataPath() + 'vehicle/')

# Initial vehicle location and orientation
initLoc = chrono.ChVector3d(0, 0, 0.5)
initRot = chrono.ChQuaterniond(1, 0, 0, 0)

# Visualization type for vehicle parts (PRIMITIVES, MESH, or NONE)
vis_type = veh.VisualizationType_MESH
# Collision type for chassis (PRIMITIVES, MESH, or NONE)
chassis_collision_type = veh.CollisionType_NONE
# Type of tire model (RIGID, TMEASY)
tire_model = veh.TireModelType_TMEASY

# Rigid terrain
# terrain_model = veh.RigidTerrain.BOX
terrainHeight = 0  # terrain height
terrainLength = 100.0  # size in X direction
terrainWidth = 100.0  # size in Y direction

# Poon chassis tracked by the camera
trackPoint = chrono.ChVector3d(-3.0, 0.0, 1.1)

# Contact method
contact_method = chrono.ChContactMethod_NSC
contact_vis = False
# Simulation step sizes
step_size = 1e-3
tire_step_size = step_size
# Time interval between two render frames
render_step_size = 1.0 / 50  # FPS = 50

# Create the MAN vehicle, set parameters, and initialize
vehicle = veh.MAN_10t()
vehicle.SetContactMethod(contact_method)
vehicle.SetChassisCollisionType(chassis_collision_type)
vehicle.SetChassisFixed(False)
vehicle.SetInitPosition(chrono.ChCoordsysd(initLoc, initRot))
vehicle.SetTireType(tire_model)
vehicle.SetTireStepSize(tire_step_size)
vehicle.Initialize()
vehicle.SetChassisVisualizationType(vis_type)
vehicle.SetSuspensionVisualizationType(vis_type)
vehicle.SetSteeringVisualizationType(vis_type)
vehicle.SetWheelVisualizationType(vis_type)
vehicle.SetTireVisualizationType(vis_type)

vehicle.GetSystem().SetCollisionSystemType(chrono.ChCollisionSystem.Type_BULLET)

# randomly create 5 boxes in the vehicle's simulation area
for i in range(5):
    side = np.random.uniform(0.5, 1.5)
    box = chrono.ChBodyEasyBox(side, side, side, 1000, True, False)
    box.SetPos(chrono.ChVector3d(np.random.uniform(-5, 15), np.random.uniform(2, 5), side / 2))
    box.SetFixed(True)
    box.GetVisualShape(0).SetTexture(chrono.GetChronoDataFile("textures/blue.png"))
    vehicle.GetSystem().Add(box)

# Create the terrain
patch_mat = chrono.ChContactMaterialNSC()
patch_mat.SetFriction(0.9)
patch_mat.SetRestitution(0.01)
terrain = veh.RigidTerrain(vehicle.GetSystem())
patch = terrain.AddPatch(patch_mat, chrono.ChCoordsysd(chrono.ChVector3d(0, 0, 0), chrono.QUNIT), terrainLength, terrainWidth)
patch.SetTexture(veh.GetDataFile("terrain/textures/grass.jpg"), 200, 200)
patch.SetColor(chrono.ChColor(0.8, 0.8, 0.5))
terrain.Initialize()

# Create the vehicle Irrlicht interface

vis = veh.ChWheeledVehicleVisualSystemIrrlicht()
vis.SetWindowTitle('MAN 10t Demo')
vis.SetWindowSize(1280, 1024)
vis.SetChaseCamera(trackPoint, 15.0, 0.5)
vis.Initialize()
vis.AddLogo(chrono.GetChronoDataFile('logo_pychrono_alpha.png'))
vis.AddLightDirectional()
vis.AddSkyBox()
vis.AttachVehicle(vehicle.GetVehicle())

# create a sensor manager
manager = sens.ChSensorManager(vehicle.GetSystem())
# create a lidar sensor
# Create the lidar sensor and attach it to the rover
offset_pose = chrono.ChFramed(
    chrono.ChVector3d(2.0, 0, 1), chrono.QuatFromAngleAxis(0, chrono.ChVector3d(0, 1, 0))
)
# lidar related parameters
# Update rate in Hz
update_rate = 5.0
# Number of horizontal and vertical samples
horizontal_samples = 800
vertical_samples = 300
# Horizontal and vertical field of view (radians)
horizontal_fov = 2 * chrono.CH_PI  # 360 degrees
max_vert_angle = chrono.CH_PI / 12
min_vert_angle = -chrono.CH_PI / 6
# Lag time
lag = 0
	\end{lstlisting}
			}   
		\end{minipage}
	}
	\caption{The generated code of finetuned  for the third round modifications of the PyChrono MAN truck simulation.(Part 1 of 2)}
	\label{fig:sft-gpt4o-mini-man-truck-3-1}
    \end{figure*}
\begin{figure*}[ht]
	\centering
	\ovalbox{
		\begin{minipage}{1.0\linewidth}
			{\tiny
				\begin{lstlisting}[breaklines=true, numbers=none]
# Collection window for the lidar
collection_time = 1. / update_rate  # typically 1/update rate
# Radius of samples to use, 1->1 sample, 2->9 samples, 3->25 samples...
sample_radius = 2
# 3mm radius (as cited by velodyne)
divergence_angle = 0.003
# Lidar return mode
return_mode = sens.LidarReturnMode_STRONGEST_RETURN
lidar = sens.ChLidarSensor(
    vehicle.GetChassis().GetBody(),  # Body lidar is attached to
    update_rate,  # Scanning rate in Hz
    offset_pose,  # Offset pose
    horizontal_samples,  # Number of horizontal samples
    vertical_samples,  # Number of vertical channels
    horizontal_fov,  # Horizontal field of view
    max_vert_angle,  # Maximum vertical field of view
    min_vert_angle,  # Minimum vertical field of view
    100.0,  # Maximum lidar range
    sens.LidarBeamShape_RECTANGULAR,  # Shape of the lidar beam
    sample_radius,  # Sample radius
    divergence_angle,  # Divergence angle
    divergence_angle,  # Divergence angle (again, typically same value)
    return_mode  # Return mode for the lidar
)
lidar.SetName("Lidar Sensor")
lidar.SetLag(lag)
lidar.SetCollectionWindow(collection_time)
lidar.PushFilter(sens.ChFilterDIAccess())
lidar.PushFilter(sens.ChFilterPCfromDepth())
lidar.PushFilter(sens.ChFilterVisualizePointCloud(640, 480, 1.0, "Lidar Point Cloud"))
manager.AddSensor(lidar)

# Create the driver system
driver = veh.ChInteractiveDriverIRR(vis)

# Set the time response for steering and throttle keyboard inputs.
steering_time = 1.0  # time to go from 0 to +1 (or from 0 to -1)
throttle_time = 1.0  # time to go from 0 to +1
braking_time = 0.3  # time to go from 0 to +1
driver.SetSteeringDelta(render_step_size / steering_time)
driver.SetThrottleDelta(render_step_size / throttle_time)
driver.SetBrakingDelta(render_step_size / braking_time)

driver.Initialize()

# output vehicle mass
print("VEHICLE MASS: ", vehicle.GetVehicle().GetMass())
# Number of simulation steps between miscellaneous events
render_steps = math.ceil(render_step_size / step_size)

# Initialize simulation frame counter
realtime_timer = chrono.ChRealtimeStepTimer()
step_number = 0
render_frame = 0

while vis.Run():
    time = vehicle.GetSystem().GetChTime()
    # Render scene and output POV-Ray data
    if (step_number % render_steps == 0):
        vis.BeginScene()
        vis.Render()
        vis.EndScene()
        render_frame += 1
    # Get driver inputs
    driver_inputs = driver.GetInputs()

    # Update modules (process inputs from other modules)
    driver.Synchronize(time)
    terrain.Synchronize(time)
    vehicle.Synchronize(time, driver_inputs, terrain)
    vis.Synchronize(time, driver_inputs)

    # update sensor manager
    manager.Update()
    # Advance simulation for one timestep for all modules
    driver.Advance(step_size)
    terrain.Advance(step_size)
    vehicle.Advance(step_size)
    vis.Advance(step_size)

    # Increment frame number
    step_number += 1
    # Spin in place for real time to catch up
    realtime_timer.Spin(step_size)
	\end{lstlisting}
			}   
		\end{minipage}
	}
	\caption{The generated code of finetuned GPT4o-mini for the third round modifications of the PyChrono MAN truck simulation.(Part 2 of 2)}
	\label{fig:sft-gpt4o-mini-man-truck-3-2}
    \end{figure*}

\FloatBarrier

\section{Limitations}
The ChronoLLM-produced PyChrono simulation scripts are rarely perfect, but are often a very useful starting point for further ``polishing'' by the user. Getting a fast start is helpful, since the initial steps towards generating a simulation script are often the most intimidating and time-consuming. While ChronoLLM provides a productivity  boost, it has several limitations:
\begin{itemize}
    \item Harmful and unpredictable content: Although ChronoLLM can reject unethical queries, these models may still generate harmful content that is misaligned with human preferences and values. This issue may arise from biases in the training data or the models' inability to discern appropriate outputs in certain contexts. Our fine-tuning process might exacerbate this issue.
    \item Insufficient domain data (in this case, multi-body dynamics related): To produce a PyChrono script from natural language, an LLM needs more than just general natural language processing capabilities. It requires foundational knowledge in mechanical engineering, including geometry, kinematics, statics, and dynamics. Training an LLM solely on the Chrono documentation and examples is brittle. There is substantial room for improvement of ChronoLLM by collecting training data that enhance's the LLM's  understanding of the underlying mechanical problem being considered, which requires a deeper domain-specific understanding (in our case multi-body dynamics).
    \item Lack of robustness: The trained models may exhibit brittleness in some situations, producing inconsistent or nonsensical outputs when faced with adversarial inputs or rare language phenomena. In other words, it is easy to trick ChronoLLM into generating incorrect code.
    \item Scalability and efficiency: Although we applied LoRA and quantization to make the model more accessible to a broader community, when combined with the original LLaMA, the models' large size and complexity can lead to difficulties in deployment, especially for users with limited computational resources. This issue may hinder the accessibility and widespread adoption of ChronoLLM or similar domain-specific LLMs.
\end{itemize}

\section{Conclusion and Future Work}
\label{sec:Conclusion}
This contribution outlined the process of training a pretrained large language model to produce ChronoLLM, a model capable of generating PyChrono simulation scripts that set up digital twins of mechanical systems. These scripts can be subsequently run in the Chrono simulator through its Python API. Our testing showed that fine-tuned models, such as \textit{gpt-40-mini-f9-t0.1}, outperformed the pretrained LLM and in-context learning approaches, achieving higher scores in metrics like BLEU, CodeBLEU, and pass@k. These results demonstrate the effectiveness of fine-tuning in enhancing functional and semantic accuracy.

To ensure fair evaluation, we introduced the ``Hacked SimBench'' baseline, which highlighted the fine-tuned models' superiority under identical training data conditions. Additionally, the use of diverse test environments validated the models' ability to generalize across various PyChrono tasks. The overall conclusion is that ChronoLLM is an effective domain-specific tool for digital twin generation. While there is plenty of room for improvement, ChronoLLM is already capable of generating scripts for a range of virtual experiments. Although PyChrono scripts are rarely perfect, ChronoLLM can provide a productivity boost in the use of Chrono.

The following are directions for future ChronoLLM improvements:
\begin{enumerate}
	\item Unlearning: As it quickly became apparent that the base LLMs picked for fine-tunning had factored in stale and therefore wrong PyChrono information, we tried to overwrite it with contemporary information associated with the latest version of Project Chrono. However, it's still possible for the old wrong information to resurface in ChronoLLM inference tasks. A possible way to fix this is called ``unlearning'' \cite{yao2024LLMunlearning}, which seeks to cause LLMs to forget the wrong, improper information. This approach remains to be tested.
	\item Multi-Modal LLMs: Recognizing the growing importance of digital twins in modern engineering, we plan to also develop multi-modal LLMs capable of processing images and videos to construct precise digital twins. For instance, providing a series of images of the mechanical system of interest is bound to help the reasoning process, since it helps clarify the ``mechanical'' part of the problem.
	\item Enhanced Tool Interaction: We aim to advance the integration capabilities of LLMs, enabling them to interact seamlessly with compilers and leverage external numerical computing resources. This initiative will focus on developing interfaces that allow LLMs to dynamically interact with a variety of software tools, enhancing their utility and efficiency in real-world applications. This enhanced interaction promises to streamline the development process, reduce error rates, and accelerate the transition from code generation to deployment. 
	\item Multi-Level Agent LLMs: Building on the success of multi-agent systems in robotics, we plan to design multi-level agent LLMs specifically tailored for the construction of simulation code. These systems will employ hierarchical decision-making structures, where agents at different levels manage specific aspects of the simulation framework, from low-level numerical computations to high-level scenario planning. This approach will facilitate more complex, adaptive, and robust simulation environments, mirroring the collaborative dynamics found in intelligent autonomous systems.
\end{enumerate}

From a high vantage point, the LLM customization process is involved, and presently comes short of the ultimate goal of having ChronoLLM generate, without any mistake, an entire digital twin. Despite its shortcomings, ChronoLLM can provide a significant help in producing a PyChrono model. Looking forward, this group is committed, in a parallel effort, to maintaining a benchmark that documents the progress of ChronoLLM over time. Ultimately, the progress of ChronoLLM is a proxy for the progress of two aspects involved in its development: the progress of the open- and closed-source LLMs that anchor ChronoLLM; and the progress of the refining and customization techniques discussed herein, which critically rely on the quantity and quality of the training data.

\section*{Supplementary information}
All open-source post-trained models discussed in this contribution are available on Hugging Face upon request. The APIs of the post-trained GPT models discussed in this paper are available upon request. In addition, all generated LLM trajectories and the evaluation infrastructure that can reproduce the results of the paper have been open-sourced in the SimBench GitHub repository.

\section*{Acknowledgments}
Partial support for the work reported herein comes from National Science Foundation project CMMI2153855.

\section*{Declarations}
All codes, data, and models used in this study will be open-sourced on GitHub. The link will be provided in the final version of the paper.


%
%

\bibliographystyle{IEEEtran}
\bibliography{BibFiles/refsChronoSpecific,BibFiles/refsMBS,BibFiles/refsSBELspecific,BibFiles/refsML-AI}   
\begin{appendices}
    \section{Prompts and Data Synthesis Pipeline}
    \begin{figure*}[ht]
	\centering
	\ovalbox{
		\begin{minipage}{1.0\linewidth}
			{\tiny
				\begin{lstlisting}[breaklines=true, numbers=none]
 You are a PyChrono expert tasked with evaluating a simulation script by comparing it against a reference script generated by experts. Your evaluation should consider both the accuracy of the script compared to the reference and adherence to best practices as outlined in the PyChrono API documentation.
					
 Here is the PyChrono code you need to evaluate:
 [The Start of Assistant's Answer]
 {code}
 [The End of Assistant's Answer]
					
 Here is the expert-generated reference code:
 [The Start of Reference Answer]
 {reference_code}
 [The End of Reference Answer]
					
 Use the following evaluation criteria and guidelines for point deduction:
  1. **Completeness (40 points total)**
   - Compare the provided code to the reference script. Deduct **15 points** for each missing essential component (e.g., system initialization, body creation, visualization) that is present in the reference script.
   - Deduct **10 points** if a component is present but lacks important details or is incorrectly configured compared to the reference.
   - Deduct **5 points** for minor omissions or slight deviations from the reference script.
  2. **Correctness (30 points total)**
   - Compare the code to the reference script. Deduct **15 points** for each incorrect use of a PyChrono API call that could lead to a change in simulation behavior.
   - Deduct **10 points** for logical errors in the code, such as incorrect joint initialization or wrong setting of body properties, especially if the reference script does it correctly.
   - Deduct **5 points** for minor inaccuracies or unnecessary API calls that deviate from the reference script.
  3. **Code Quality (10 points total)**
   - Evaluate the readability, structure, and documentation of the code against the reference script. Deduct **5 to 10 points** for poor readability, structure, or lack of meaningful variable names and formatting.
   - Deduct **5 points** for insufficient comments or failure to follow documentation best practices, especially if the reference script provides better documentation.
  4. **Efficiency (10 points total)**
   - Evaluate the efficiency of the code compared to the reference script. Deduct **5 points** for each instance of unnecessary calculations, redundant code, or inefficient use of APIs that is optimized in the reference script.
   - Deduct **3 points** for missing obvious optimization opportunities that the reference script implements.
  5. **Error Handling and Robustness (5 points total)**
   - Assess the error handling and robustness of the code. Deduct **5 points** for lack of basic error handling or failure to account for common issues that the reference script handles.
   - Deduct **3 points** for inadequate handling of edge cases compared to the reference script. 
  6. **Use of Visualization Tools (5 points total)**
   - Compare the use of visualization tools in the provided code to the reference script. Deduct **3 to 5 points** for incorrect or inadequate visualization setup as per the reference script.
   - Deduct **2 points** for minor visualization issues, such as suboptimal lighting or incomplete setup of visual elements, compared to the reference.
					
 Avoid position biases and ensure that the order in which the responses are presented does not influence your decision. Do not allow the length of the responses to influence your evaluation. Do not favor certain names of the assistants. Be as objective as possible.
					
 After providing your explanation, output the final score using the format '[[x]]' where x is the score assigned to the assistant's answer.
					
 Reference the PyChrono API documentation provided here: {api_documentation_link}
					
 Provide the evaluated score and a brief explanation of the deductions below:
				\end{lstlisting}
			}
		\end{minipage}
	}
	\caption{Instructions for using J-LLM with expert written API documentation and ground truth code.}
	\label{fig:eval_framework}
\end{figure*}

\begin{figure*}[ht]
	\centering
	\ovalbox{
		\begin{minipage}{1.0\linewidth}
			{\tiny
				\begin{lstlisting}[breaklines=true, numbers=none]
Your task is to generate high-quality, context-rich question-and-answer pairs based on the provided PyChrono code.
				
Imagine you are a **normal PyChrono user** encountering code from another language model. To resolve your issue, you must provide very concrete context when asking for help from another LLM.
				
Requirements:
- Generate {num_pairs} Q&A pairs that are clear, specific, and focused on key points from the context.
- Each question must be understandable without prior knowledge, including necessary background details from the context.
- Avoid vague references like "in this simulation" or "in the code", provide full context in the question.
				
Instructions:
1. **Question Generation**: Write a concise, clear question tied to a specific aspect of the code, ensuring enough context is given without revealing the answer. Imagine you're explaining the problem as a typical PyChrono user, needing to convey the full problem to get help from another LLM.
2. **Answer Generation**: Provide a precise answer based only on the provided context. Include explanations, corrections (if applicable), and detailed reasoning where necessary.
				
Output Format:
Generate the output in JSON format as follows:
{{
	"instruction": "<clear, context-rich question from the perspective of a *normal PyChrono user*, with full problem context>",
	"input": "",
	"output": "<detailed answer with explanations and code corrections (if needed)>"
}}
				
Example:
{{
	"instruction": "How can you set the collision system type in PyChrono to use the Bullet physics engine?",
	"input": "",
	"output": "To set the collision system type in PyChrono to Bullet, use `GetSystem()` to access the system and `SetCollisionSystemType()` to set it: `vehicle.GetSystem().SetCollisionSystemType(chrono.ChCollisionSystem.Type_BULLET)`."
}}
				
Ensure all questions are **specific** and **contextually rich**, with **no vague references** like "in the code". Provide clear, detailed answers that fully explain the issue and solution.
				
Context:
{markdown_content}
				\end{lstlisting}
			}
		\end{minipage}
	}
	\caption{Instructions for using LLM to sanitize data.}
	\label{fig:sani1}
\end{figure*}

\begin{figure*}[ht]
	\centering
	\ovalbox{	
		\begin{minipage}{1.0\linewidth}
			{\tiny
				\begin{lstlisting}[breaklines=true, numbers=none]
You're an expert in the PyChrono simulator. Your task is to generate question-and-answer pairs for the given context using the format below. 
When generating the question, imagine you are a user of the PyChrono simulation but do not have access to the given context. So DO NOT use phrases like 'in the script' or 'in the simulation' without given any context!!! 
Instead, ask about the problem with clear and detailed context directly!
For generating the answer, base it on the given text and be as detailed as possible.
					
Requirements:
- Create {num_pairs} question-and-answer pairs.
- Avoid phrases like 'in the script' or 'in the simulation' without given any context, be specific!
- The questions should be diverse, reflecting different ways users might phrase them. 
- Include best practices or common mistakes in the explanations where appropriate.
					
Format:
{{
	"instruction": "<diverse question, may include code>",
	"input": "<use input only when necessary>",
	"output": "<detailed explanation, may include code>"
}}
					
Markdown Content:
{markdown_content}
				\end{lstlisting}
			}
		\end{minipage}
	}
	\caption{Instructions for using LLM to sanitize data.}
	\label{fig:sani2}
\end{figure*}

\begin{figure*}[ht]
	\centering
	\ovalbox{
		\begin{minipage}{1.0\linewidth}
			{\tiny
				\begin{lstlisting}[breaklines=true, numbers=none]
Your task is to generate question-and-answer pairs for a given PyChrono markdown file, focusing on **debugging tasks**. Follow the detailed instructions below for creating these pairs.
					
Requirements:
- Generate {num_pairs} pairs of questions related to **incorrect code** and provide detailed answers that describe the issues and how to fix them.
- Each question should focus on debugging or fixing errors in the given PyChrono code.
- The incorrect code must appear in the **instruction** section, framed as if a normal **PyChrono user** is asking for help with a bug or issue.
- The answer must describe what the error is and provide the corrected version of the code, along with a natural language explanation.
					
Rules for Bug Code Generation:
					
1. **API Misuse**: Common mistakes involve calling methods in the wrong order or using an API in a context where it's not valid.
	- **Example Error**: Calling `system.SetStep(0.01)` before the system is initialized.
					
2. **Misspelled API Names**: Simple typos or incorrect capitalization that lead to errors.
	- **Example Error**: Typing `ChSystemNSC` as `ChSystemNCS`; using 'chrono.ChVector3d' as 'chrono.ChVector3D'; using 'chrono.ChFramed(...)' as 'chrono.ChFrame(...)'.
					
3. **Wrong Parameter Types**: Passing values that are not valid for specific function parameters.
	- **Example Error**: Passing a string (`"1.0"`) instead of a float (`1.0`) to `SetMass()`.
					
4. **Incorrect Initialization**: Leaving out necessary steps like setting the mass or inertia of an object, leading to runtime issues.
	- **Example Error**: Failing to set the position of a `ChBody()`.
					
5. **Logic Errors**: Errors that don't crash the code but lead to unrealistic simulation behavior.
	- **Example Error**: Setting all joints as fixed, making the system immovable.
					
6. **Wrong Data Types or Values**: Using nonsensical values that make the simulation physically incorrect.
	- **Example Error**: Setting mass to a negative number.
					
7. **Unreasonable Time Step**: Setting a time step so large that it causes instability in simulations.
	- **Example Error**: `system.SetStep(1.0)` instead of a reasonable value like `0.01`.
					
Instructions:
1. Imagine you are a **normal PyChrono user** who has encountered a bug or error in the code and is seeking help **without knowing the exact issue**.
Again, the user should not know the exact error but should provide the code snippet that is causing the problem, framed as a question!!!
					
2. The incorrect code should appear in the **instruction** section of the output, and the **input** section should remain empty.
3. Provide a detailed natural language explanation of what the error is, why it occurs, and how to fix it, including the corrected version of the code.
4. Generate a diverse set of questions focusing on different types of errors and issues in the PyChrono code.
5. The code provided in the markdown file is ***absolutely correct***, and it should be used as the basis for generating the Q&A pairs.
					
Output Format:
Generate the output in JSON format as follows:
					
{{
	"instruction": "<incorrect code, framed as a user question>",
	"input": "",
	"output": "<detailed explanation of the error and the corrected code>"
}}
Example:
{{
	"instruction": "This code isn't working correctly, and I can't figure out why: ```python\n# Create Chrono system\nsystem = chrono.ChSystemNCS()\nsystem.SetCollisionSystemType(chrono.ChCollisionSystem.Type_BULLET)\nsystem.SetGravitationalAcceleration(chrono.ChVector3d(0, 0, -9.81))\n```",
	"input": "",
	"output": "The error here is a typo in the system initialization function. The correct function is `ChSystemNSC()` (Non-Smooth Contacts), not `ChSystemNCS()`. Here's the corrected code:\n\n```python\nsystem = chrono.ChSystemNSC()\nsystem.SetCollisionSystemType(chrono.ChCollisionSystem.Type_BULLET)\nsystem.SetGravitationalAcceleration(chrono.ChVector3d(0, 0, -9.81))\n```\nThe corrected code now initializes the system properly for non-smooth contacts."
}}
PyChrono Script Markdown file Content:
{markdown_content}
				\end{lstlisting}
			}
		\end{minipage}
	}
	\caption{Instructions for using LLM to sanitize SFT data.}
	\label{fig:sani3}
\end{figure*}

    \label{secA1}
\end{appendices}
\end{document}